# TinyML-Enabled IoT for Sustainable Precision Irrigation

Kamogelo Taueatsoala[1], Caitlyn Daniels[1], Angelina J. Ramsunar[1], Petrus Bronkhorst[2], Absalom E. Ezugwu[1,*]
[1] Unit for Data Science and Computing, North-West University, 11 Hoffman Street, Potchefstroom 2520, South Africa.
[2] Unit for Environmental Sciences and Management, North-West University, Potchefstroom, South Africa

Corresponding author email: Absalom E. Ezugwu (Absalom.Ezugwu@nwu.ac.za)

**Abstract.** Small-scale farming communities are disproportionately affected by water scarcity, erratic climate patterns, and a lack of access to advanced, affordable agricultural technologies. To address these challenges, this paper presents a novel, edge-first IoT framework that integrates Tiny Machine Learning (TinyML) for intelligent, offline-capable precision irrigation. The proposed four-layer architecture leverages low-cost hardware, an ESP32 microcontroller as an edge inference node and a Raspberry Pi as a local edge server, to enable autonomous decision-making without cloud dependency. The system utilizes capacitive soil moisture, temperature, humidity, pH, and ambient light sensors for environmental monitoring. A rigorous comparative analysis of ensemble models identified Gradient Boosting as superior, achieving an $R^2$ score of 0.9973 and a Mean Absolute Percentage Error (MAPE) of 0.99%, outperforming a Random Forest model ($R^2$ = 0.9916, MAPE = 1.81%). This optimized model was converted and deployed as a lightweight TinyML inference engine on the ESP32, predicts irrigation needs with exceptional accuracy (MAPE < 1%). Local communication is facilitated by an MQTT-based LAN protocol, ensuring reliable operation in areas with limited or no internet connectivity. Experimental validation in a controlled environment demonstrated a significant reduction in water usage compared to traditional methods, while the system's low-power design and offline functionality confirm its viability for sustainable, scalable deployment in resource-constrained rural settings. This work provides a practical, cost-effective blueprint for bridging the technological divide in agriculture and enhancing water-use efficiency through on-device artificial intelligence.

## 1. Introduction

The discussion in global communities on how to address the problem of food security focuses more on large-scale farmers as the answer. However, this is not an inclusive strategy as food insecurity remains a prevailing challenge in rural communities, where less attention is given to the various innovative and technological advancements for sustainable food production (van den Berg & Walsh, 2023). Small-scale or subsistence farmers play major roles in solving food insecurity at a household level. By equipping small-scale farmers with suitable and low-cost smart technologies such as Internet of Things (IoT) irrigation systems, which react to the real-time environmental conditions, subsistence agriculture can be turned into a more data-driven and sustainable practice (Kapari et al., 2023; Kazi, 2025). With the integration of environmental monitoring and machine learning capabilities in the design, the technology assists small-scale farmers to maximize water usage, predict crop productivity based on environmental factors, and react to weather fluctuations (Oluwatayo, 2019). In the era of uncertain climatic conditions, the technology enhances food productivity at a lower price, and food security at the household level becomes a reality.

Climate change is seen as an increase in environmental temperatures, as well as low rainfall with erratic patterns. In developing countries, about 70 percent of people living in rural areas depend on small-scale farming, which has been associated with low productivity and unstable income because of the effects of climate change on rainfall (Vermeulen et al., 2010). According to the World Commission on Environment and Development (WCED), approximately 80 countries that account for 40% of the world's population are already suffering from water scarcity (Hamdy et al., 2003). These changes are more impactful in rural areas whose livelihoods are dependent on natural resources (Rankoana, 2016). According to Nhamo and Chilonda (2012) and Matimolane and Mathivha (2025), water scarcity, which is a result of unreliable rainfall, poses a challenge for small-scale farms in Africa. Crop yields produced by rainfed agriculture declined by 50 percent before the year 2020, which increased the food insecurity of vulnerable groups (Food & Nations, 2010).

In 2015, the United Nations (UN) issued a plan called the U.N. 2030 Agenda. The plan had a set of seventeen Sustainable Development Goals (SDGs) as well as a series of 126 targets that were required to be reached by 2030. Within all of that, they specified the need to reduce poverty and end world hunger, which are Sustainable

Development Goal 1 (SDG 1) and SDG 2, respectively (Davarpanah et al., 2023). Increasing food production in a country is undoubtedly the most effective way to reduce hunger and poverty (Tsoukas et al., 2022). The attainment of these goals has been delayed because of the effects of climate change on agriculture, as well as the lack of modernisation of agriculture in developing countries. The water scarcity significantly affects productivity and threatens the sustainability of small-scale farming, which is vital for food security in rural communities. The shortage has led the arid and semi-arid countries to rely more on food imports since the local agricultural sector isn't able to meet the food demand in the country, which further increases the prices of the food, leading to exacerbated levels of food insecurity and poverty (Hamdy et al., 2003; Jain et al., 2024). Small-scale agriculture is an important part of achieving SDG 2; small-scale farmers can end hunger in communities, thereby promoting sustainable agricultural practices.

The traditional irrigation methods are not energy-efficient enough for sustainable food production among small-scale farmers (Salman et al., 2025). For instance, the traditional irrigation systems do not consider the prevailing environmental conditions, crop water demand, soil moisture, and weather prediction. The hard-coded operations of the existing irrigation systems are not compliant with crop demands in small-scale farming environments. In addition, dependence on cloud computing and the remote monitoring strategy of many available solutions requires high-speed Internet connectivity. This service is not available or reliable in rural locations that practice small-scale farming, limiting the adoption of artificial intelligence in scaling the irrigation process.

The adoption of Artificial Intelligence (AI) has become the leading solution to achieving the SDGs, but the implementation in the rural context has been restricted because of the digital divide. AI-based irrigation systems are expensive, which increases the inability of small-scale farmers to have access to the technology. The inaccessibility of the technology increases inefficiencies in their farming practices and limits productivity. All these limitations put small-scale farmers at a disadvantage compared to large commercial farmers in terms of productivity and market presence Arakpogun et al. (2020). This shows that there is a high demand for scalable and affordable technology to be integrated into small-scale farms for water management. With the growing population, the agricultural sector needs to become smart as well as accustomed to the different problems experienced by small-scale farmers in rural areas (Mitra et al., 2022). Lightweight machine learning models can be harnessed to achieve smart and precise irrigation in small-scale farming communities.

TinyML has the potential to bridge the gap between small-scale and commercial farmers by enabling real-time, low-power data processing on affordable edge devices (Warden & Situnayake, 2019). While the technology facilitates smart agriculture in low-resource environments, it is helpful to disadvantage farmers in increasing their crop yields and promotes food sustainability at a reasonable price. This study opines that implementing TinyML-driven precision irrigation systems can improve agricultural productivity and sustainability, leading to long-term food security in resource-constrained regions. It maximizes the irrigation times and resource distribution through AI-based predictive models. While the study aims to establish a foundation for developing a context-sensitive, technology-driven solution that supports sustainable agricultural practices in low-resource settings, the study is structured around several key research questions, including:

i. How can IoT architecture be effectively designed for real-time smart irrigation in resource-constrained environments?
ii. Which communication protocols best ensure reliable, low-latency data transmission in areas with limited connectivity?
iii. What strategies can ensure the system is cost-effective, scalable, and adaptable while maintaining functionality comparable to commercial systems?
iv. What data preprocessing techniques are most effective for preparing noisy sensor data for TinyML model training?
v. How can edge computing be leveraged to minimize latency and enhance real-time processing?

In answering the stated research questions, the following specific scientific contributions have been made in this study:

- Design and implementation of a reusable, edge-first reference architecture organised into four layers: perception, edge node, edge server, and edge service, that locates decision-making on the ESP32 microcontroller, while the Raspberry Pi is constrained to telemetry, local storage, and data visualisation.
- Demonstration of a LAN-only Message Queuing Telemetry Transport (MQTT) data plane with a documented topic schema and payload contract spanning sensor telemetry, control commands, and configuration, enabling deterministic local operation without Internet dependence.
- Development of a TinyML-based predictive model for weather and sensor measurements analysis in making irrigation decisions. The model supports offline training using historical weather and irrigation data for lightweight inference on embedded devices.

These questions are designed to probe the core technical and practical challenges in developing a smart irrigation system to bridge the gap between advanced AI irrigation technologies and the real-world needs of small-scale farmers. Answering the stated research questions will not only inform small-scale farmers of the challenges or the benefits, but it will also contribute to a broader understanding of the emerging AI technologies and how they can be localised to enhance climate resilience within underprivileged communities.

The rest of the paper is organized as follows: Section 2 presents the discussion of relevant studies and other related concepts in the existing literature, while Section 3 focuses on the methodological approach in the design and implementation of the proposed framework. Section 4 presents results and discussion, and the conclusion is reported in Section 5.

## 2. Literature Review

Recent advances in the field of digital technologies have influenced the agricultural sector significantly (Smith et al., 2010). The introduction of IoT technologies in agricultural irrigation is a major deviation from conventional farming practices to precision agriculture (Padhiary et al., 2025). The transition to smart precision farming makes room to address pressing issues like food insecurity at a household level and water shortages in the agricultural sector. Despite such a positive change, the appropriate precision systems would depend on the effective use of IoT architecture and sensor integration practices that will enable real-time data collection, transmission, and actuation (Pandey et al., 2025). The successful functionality is not directly linked to the performance of each component but to the robustness of the overall structure of an architecture that facilitates integration, scalability, and flexibility to various conditions of farming (García et al., 2020).

With the introduction of IoT in the agricultural industry, the irrigation system has been revolutionized, serving as a backbone to sustainable precision farming. The IoT architecture does not have a universal architectural framework, with each specific framework potentially benefiting the implementation and integration of IoT devices. As stated in Mehmood et al. (2021), the basic three-layered architecture developed during the evolution of IoT technology remains relevant in all subsequent developments in IoT architecture. The architecture consists of a perception layer, which is equipped with sensors that will sense and gather environmental data. The second layer is the network or transmission layer that is responsible for transmitting the data gathered in the perception layer to the application layer. The Topmost layer is the application layer, which receives sensory data from the network layer and analyzes the data for decision-making. The foundation of conceptualising IoT systems and the initiation of more modular architectures lies in these layers.

Precision irrigation systems powered by IoT sensors, automated monitoring, and machine learning have the potential to improve the efficiency of water usage in farming regions that experience acute water scarcity (Abioye et al., 2022; Bwambale et al., 2022; Lakshmi et al., 2023). Authors in Kamilaris and Pitsillides (2016) presented a framework for an IoT-enabled smart agricultural system targeted at sustainable food production. The Agri-IoT model was based on a three-layered architecture, with the bottom layer serving as the devices and communication planes, the middle layer dealing with data and data analytics, and the top layer serving the applications and end-user planes. The design assumes a hierarchical approach and focuses on a distributed architecture, which is flexible, and seeks to enable the integration and interoperability of heterogeneous sensor streams.

García et al. (2020) noted that the lack of a standardised framework for IoT architecture presents an integration and interoperability challenge across sectors. The application of standardised architecture will offer standard interfaces, which facilitate standard communication, exchange of data, and simplicity of integration with various components. The standardised framework is necessary because, beyond the challenges of interoperability and integration, it adversely affects the adoption of precision systems that utilise IoT devices like the smart irrigation system. Small-scale farming particularly finds it difficult to overcome the lack of standardisation because of the heterogeneity of the devices.

In recent advancements of IoT-supported architecture, especially in automated irrigation, GS Campos et al. (2019) presented a smart and green framework of an IoT smart irrigation system. While the frameworks are standardised, it is a framework that has been established in addressing the absence of complete IoT platforms. The framework aims to offer agricultural monitoring services that are reusable and modular, and offer an ease of integration with a wide variety of IoT infrastructures. The adoption of low-cost devices (Arduino and Raspberry Pi) in our proposed framework demonstrates its usage in modular devices and provides some scaling capabilities. Insights from the studies show progress in the evolution of different theoretical IoT architectures with the introduction of interoperability and scalability issues in the absence of a standardised IoT architecture and sensor integration framework. This opens a door for future studies to target achieving a standardised IoT framework that is scalable, interoperable, cost-effective, and simple to deploy to overcome fear among small-scale farmers in adopting smart technology.

### 2.1 Sensor Data Preprocessing

Smart and precision agricultural systems will not offer any satisfactory performance if the information retrieved by sensors is not in any meaningful format for processing by a machine learning algorithm. The preprocessing of sensor data ensures that data are prepared in a meaningful and correct format before delivery to the TinyML module for further analysis in resource-constrained microcontrollers (Nguyen et al., 2025). This is significant for real-time decisions at the edge, eliminating reliance on cloud computing (Sabovic et al., 2023). In a smart irrigation system, the sensory data acquired by the perception layer is generally noisy, with the presence of outliers and anomalies. Studies in Iorliam et al. (2022) and Nawaz and Babar (2025) present preprocessing and data cleaning methods in IoT integrated systems for filtering and elimination of irregularities in sensor data. The studies further observe that outlier and anomaly detection methodologies are instrumental in improving accuracy since inconsistent values, if not effectively handled, may result in abnormal behaviour of the irrigation model, leading to inefficiency in watering decisions.

Employing methods like Min-Max in ensuring that input values scale at the same rate indicates the validity of feature engineering or normalisation methods in enhancing data compatibility and the efficiency of irrigation models. Kamilaris and Pitsillides (2016) refer to the data fusion method as one of the preprocessing methods, where sensor data generated by the same type of sensor are merged to enhance data quality. The pre-processing methods facilitate informed decision-making in the irrigation system.

As indicated in Nawaz and Babar (2025), these techniques are fundamental in ensuring the generation of meaningful sensor data for processing by a TinyML model to achieve computational and memory efficiency. To cater to the limited computational and memory abilities of microcontrollers, ESP32 and the Raspberry Pi, David et al. (2021) reported the applications of quantisation, pruning, and feature extraction methods. Quantisation is an operation that aims at minimal memory requirements in computing by denoting numbers using lower bit representations, while pruning removes unnecessary model weights using feature extraction. Bhushan et al. (2025) opined that quantisation, pruning, and compression are relevant preprocessing methods of TinyML since it significantly reduce memory consumption, the amount of energy, and accelerate inference time without losing much accuracy.

To achieve energy efficiency in TinyML-based systems, efficient data preprocessing is necessary. This presents a new layer of challenge in achieving a balance between the complexity of data preprocessing and energy consumption (Katib et al., 2025). Although TinyML is an edge-based real-time processing, it is limited in terms of computation and power availability constraints, limiting its functionality in delay-sensitive and heavy calculation environments, which undermines real-time responsiveness and energy usage. A compact IoT data processing stack is required to achieve energy-efficient on-device inference on resource-constrained devices (Sabovic et al., 2023). Efficient preprocessing of raw sensor data into a meaningful format has shown improvement in water efficiency by between 4.3 and 20.7% in the irrigation system study conducted by (GS Campos et al., 2019).

Various developments are evolving from the combination of TinyML and IoT in farming methods, highlighting steady growth in technical knowledge and practical farming experience. For example, Bhattacharya and Pandey (2024) proposed a secure and fast model that uses TinyML and blockchain to monitor soil quality. The study demonstrated the usage of such technology beyond data collection for environmental sustainability. Other related projects that incorporated TinyML in their design are reported in Kulkarni and Bhudhwale (2024) for real-time environmental monitoring, Elhanashi et al. (2024) broadened the discussion to ethical and technological challenges, Mohammed and Munir (2025) discussed the potential of these technologies in driving long-term agricultural growth.

### 2.2 Edge Computing

Edge computing is an emerging paradigm in IoT systems that allows local data processing and decision-making by shifting the computational requirements towards the edge of the network and the data source (Varghese et al., 2016). This paradigm transcends the fundamental inefficiencies of centralised cloud-based systems, particularly latency, inefficient utilisation of bandwidth, and overreliance on Internet connectivity (Mehmood et al., 2021). Edge computing is more responsive and resilient because it minimises the need to transmit data to remote servers for processing, making it a powerful technology for remote agricultural systems, such as a smart irrigation system (Zhang et al., 2025).

Systems across various domains, such as smart healthcare, smart agriculture, smart city, etc., use edge computing as a component of a multi-tier layered architecture, where each of the layers is distributed in its functionality (Nawaz & Babar, 2025). The multi-level executing plan in the design was aimed at a perception or device layer in which the distributed IoT devices and sensors are installed directly in the field to acquire raw data. The second level is the edge or processing layer that is involved in local processing of data to minimise the amount of data to be forwarded to the cloud. The execution of TinyML in this layer reduces latency and bandwidth utilization because it supports real-time decision-making on constrained edge devices in the network. The third level is typically the cloud/application layer, which provides permanent data storage and user interfaces for computing-intensive real-time field monitoring.

The widespread use of Raspberry Pi's and Arduino's across the reviewed smart irrigation systems emphasises a shift toward designing a cost-effective, open-source, and accessible smart agricultural system suitable for adoption by small-scale farmers (Dahane et al., 2022). These devices can be used strategically according to their respective strengths for different processing requirements within the system architecture (García et al., 2020). The implementation of Arduinos was noted in multiple studies in the design of embedded systems that collect environmental data and perform lightweight TinyML inference for real-time decision-making directly at the source. Raspberry Pi functions as a local edge gateway and performs further processing on data received from embedded sensor nodes in preparation for analytics and transmission to the cloud. Beyond processing and aggregation, Puranik and Mahmood (2025) utilized Raspberry Pi for on-site data storage and processing for situational reports in the monitoring regions, promoting timely interventions, instead of relying on the centralized cloud for analytics. While these technologies offer promising solutions that may bridge the existing technological divide between commercial farmers and small-scale farmers, operational challenges such as high initial costs and limited technical skills among small-scale farmers in rural areas hinder effective implementation (García et al., 2020).

While current literature surrounding edge computing in IoT applications such as smart agricultural systems shows great potential in providing cost-effective, accessible solutions for small-scale farmers in rural areas, there are significant gaps that must be addressed. Additionally, there is limited research on standardised edge computing frameworks in an agricultural context. The development of such frameworks would support better interoperability and integration across diverse IoT devices (Nawaz & Babar, 2025). There is minimal discussion on long-term sustainability because many of the reported frameworks have not been tested over multiple seasons through various climatic conditions, as they are only tested in controlled laboratories or fields. In a similar perspective, O'Grady et al. (2019) noted that many edge implementations are still cloud centric. While decision-making at the edge is efficient, data is still forwarded to the cloud for further analytics rather than performing the analytics locally. The integration of TinyML and advanced artificial technologies for edge-centric analysis is underexplored in the agricultural sector.

### 2.3 Communication Protocols

Efficient and reliable communication is necessary for seamless data communication in any smart solutions powered by IoT (Khan et al., 2025; Yiğitler et al., 2020). The smart irrigation systems, which are based on the IoT, rely on communication protocols. The protocols facilitate a smooth flow of data among nodes, edge gateways, and end-user applications. These protocols are used to regulate the communication between devices, guaranteeing the security of the data during transmission from the source to the destination (Sabovic et al., 2023). Real-time responsiveness and system autonomy are achieved based on the strength of the communication protocol, especially in resource-constrained rural settings where access to Internet connectivity is limited.

In a smart irrigation system, the communication protocols facilitate the linking of various components, transfer of data between sensors and the decision unit, and the commands between the decision unit and the actuators when controlling irrigation (Munir et al., 2021). The different communication protocols in the literature are generally classified based on the Open Systems Interconnection (OSI) or Transmission Control Protocol/Internet Protocol (TCP/IP) layer where the protocol is located. Various protocols are executed at different layers of the network architecture for reliable data transmission and energy efficiency (Popović et al., 2017).

The common wireless protocols used in the perception layer include LoRa with LoRaWAN as an extension (Long Range Wide Area Network), Zigbee, Bluetooth Low-Energy (BLE), and Wi-Fi, utilized in the multi-tier IoT architecture (Codeluppi et al., 2020; Mehmood et al., 2021; Premkumar & Sigappi, 2022). These protocols support long-distance communication, stable data transmission, low-power consumption, and ease of configuration. The protocols enable the network layer to transmit sensed data to upper layers, and they are suitable for deployment in small-scale farming regions (Abba et al., 2019).

In the application layer, communication protocols such as MQTT and Hypertext Transfer Protocol/Hypertext Transfer Protocol Secure (HTTP/ HTTPS) are commonly used (Premkumar & Sigappi, 2022; Sabovic et al., 2023). The protocol consumes low energy and possesses the ability for reliable data transmission from the source to the destination. While MQTT is a widely recognised protocol for energy efficiency and low overhead, there are minimal reports of its adoption in real-life projects. Factors that influence communication efficiency include a lack of a universally standardised architecture and time synchronisation, hindering interoperability in applications that involve heterogeneous components (Matekaire, 2021).

## 3. Methodology

The methodology of this research is structured to systematically achieve the study's objective of developing a robust and accurate model for predicting irrigation needs. It adheres to a standard research process, comprising four main sequential phases: (1) system design and data collection, (2) experimental data preparation and feature engineering, (3) model development and comparison, and (4) rigorous validation and evaluation. This structured approach ensures the research is replicable, its findings are valid, and the final model is suitable for real-world deployment in a precision irrigation system. The first phase, system design and data collection, establishes the empirical foundation. It details the conceptual architecture of the proposed IoT-enabled smart irrigation system, specifying the types and roles of sensors (soil moisture, ambient light, pH, temperature, humidity) for environmental monitoring. The procedure for data acquisition, including the duration, frequency, and conditions of sensor readings that yielded the final dataset of 30,001 samples, is explicitly defined to ensure transparency and reproducibility.

The experimental data preparation and feature engineering phase transforms raw sensor data into a format suitable for machine learning. This involves initial exploratory data analysis (EDA), including the generation of descriptive statistics (e.g., Table 1) and correlation matrices (Fig. 1) to understand data distributions and inter-variable relationships. Subsequently, domain-specific feature engineering is applied to create agriculturally meaningful predictors from the five base sensor readings. This results in an expanded feature set that includes derived metrics such as moisture deficit, environmental stress index, and categorical light intensity levels, enhancing the model's ability to capture complex agro-environmental dynamics. The core model development and comparison phase employs a controlled experiment to identify the optimal predictive algorithm. Two state-of-the-art ensemble learning models, Random Forest and Gradient Boosting, are implemented and trained on an identical, partitioned dataset (80% for training, 20% for testing). Their performance is rigorously compared using a comprehensive suite of metrics, including $\boldsymbol{R^2}$, RMSE, MAE, MAPE, and computational efficiency (training/inference time), as detailed in the comparative results (Table 2). This direct comparison under standardized conditions provides empirical evidence for model selection.

Finally, the validation and evaluation phase subjects the selected model to stringent assessment to confirm its reliability and readiness for deployment. This includes analysis of residuals and error distributions to verify model assumptions (e.g., normality of errors), learning curve analysis to diagnose bias-variance trade-offs and overfitting, and statistical significance testing (e.g., paired t-tests) to confirm the superiority of one model over another. The outcome is a validated, high-performance model alongside clear deployment specifications, completing a methodological pipeline from system conception to a validated analytical artifact ready for integration into an intelligent irrigation system.

### 3.1 Prototype System Architecture Desing

The prototype development of the proposed smart irrigation system started with hardware assembling and calibration of hardware and software components. The ESP32 firmware was created using the Arduino framework with simple rule-based irrigation logic to test the functionality of all components. A TinyML predictive model is trained offline using historical weather and irrigation data for subsequent analyses of the sensor readings. The model predicts when it is necessary to irrigate based on the periodic environmental records and weather data. The microcontroller initiates the watering process by enabling the relay to switch on the pump based on the outcome of the data analysis. To evaluate the inputs and decisions, a rudimentary logging system is integrated to record them inputs and decisions. Sensors and irrigation records are stored continuously for future analytics. A small NoSQL database and lightweight databases (TinyDB) are deployed to the Raspberry Pi, allowing it to perform efficient read/writes with limited overhead. TinyDB fits the low-resource constraints of the system and eases the task of managing data in the embedded platform. The lightweight web interface allows users to view the status of the system, sensor data, and the history of irrigation by accessing the ESP32 access point.

The initial development cycle referred to the baseline communication and control logic, focused on hardware and actuator control logic validation. This cycle aimed to ensure that sensor data was correctly read by the ESP32 microcontroller, transmitted through the system architecture, and acted upon by the actuator unit to activate irrigation cycles. This cycle includes sensor calibration to distinguish various soil conditions and testing under controlled conditions to measure and classify thresholds. The calibration values were set at dry air (no contact with soil) to establish a baseline value for completely dry = 3699; dry soil = 3500 and 3699; wet soil to capture post irrigation conditions = 1500 and 2900. The identified thresholds informed the logic of early rule-based training. After sensors' calibration, the ESP32 firmware collects sensor readings and tests the collected values against the pre-determined threshold to trigger rule-based irrigation. The second cycle follows the principle of on-device machine learning to make autonomous irrigation decisions. This cycle aims at incorporating the trained TinyML model in the ESP32 firmware and pre-processing the sensor data.

Sensor data preprocessing logic was implemented to ensure a clean and consistent format of the data samples. Techniques like sensor reading smoothing using a moving average filter were applied to reduce the effect of sudden fluctuations. Normalisation was used to scale soil moisture readings to a [0,1] range to align with the model's input configuration. A pretrained model was converted to TensorFlow Lite for Microcontrollers and embedded into the ESP32's firmware to classify monitored soil conditions. The deployed model was tested in simulated conditions to validate the functionality of the inference model. Following the deployment of the TinyML inference model and the validation of its decision-making capabilities in the second cycle, the dashboard integration is implemented for data logging and real-time visualisation through a web-based platform. The goal of the web-based dashboard is to enable small-scale farmers to observe, monitor, and validate system behaviour without high-level technical expertise and consistent Internet connectivity.

As depicted in Figure 1, the proposed system relies on low-cost automation using IoT sensors, ESP32, TinyML model, and Raspberry Pi deployed across the layers of the system. The four layers of the architecture are the perception layer, edge node layer, edge server layer, and edge service layer, allowing sensor integration, sensor data preprocessing for machine learning analytics, edge computing, local data storage, and communication protocols. The perception layer is responsible for sensing the environment, and irrigation actuators are placed in this layer. The edge node layer acquires the environmental data, preprocesses it, and makes inferences based on the ESP32 microcontroller. The edge server layer handles the storage request and local hosting for virtualization of the data analysis outcomes, while the edge service layer allows farmers to monitor their fields using the local web dashboard.

The working principle of the design allows continuous monitoring of the soil state and atmospheric conditions in deciding the optimal irrigation level for more efficient water resource management and storing of the data for further analysis. This helps reduce water wastage, minimising intensive labour, and eliminating the element of guesswork in the scheduling of irrigation.

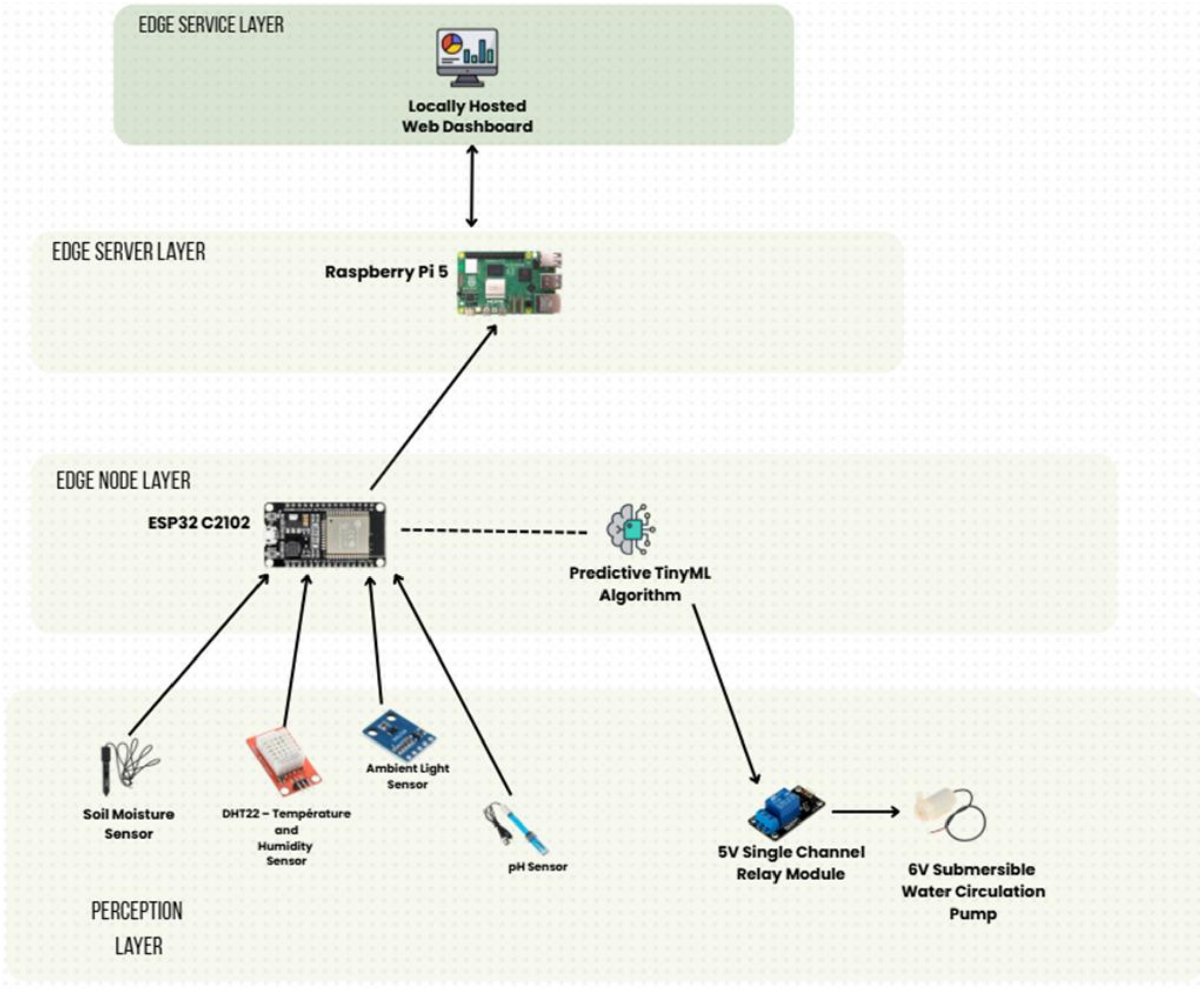

Figure 1: Overview of the System Architecture

The hardware components considered in the design of the proposed framework are in line with the study's objective to provide cost-effective, sustainable, and effective solutions that address the challenges faced by small-scale farmers. Table 1 presents the hardware components and the description of their functionalities.

Table 1: Hardware Components Descriptions

| **Hardware Component** | **Description/Function** |
|---|---|
| 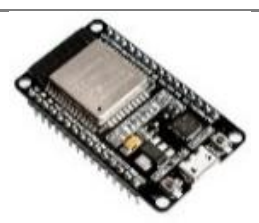 | *ESP32* -Functions as the edge node, collecting sensor data and executing local actuation decisions. |
| 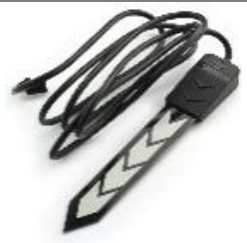 | *Capacitive soil moisture sensor* – Measure soil moisture content. |
| 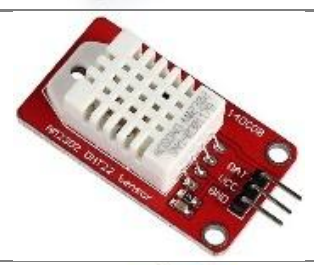 | *DHT22 – Temperature and humidity sensor* - Measures ambient temperature and humidity as it influences soil evaporation rates and crop water needs. |
| 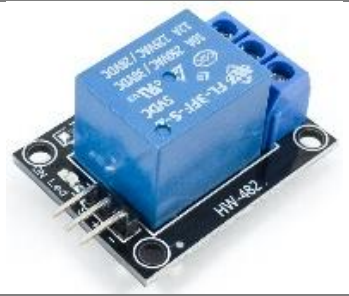 | *5V single channel relay module* - Controls the submersible water pump based on the decision made by the TinyML model. |
| 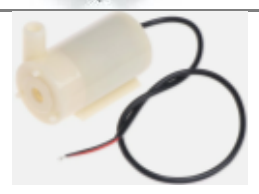 | *6V submersible water pump* – Pumps water containing nutrient solution from the reservoir to the plants. |
| 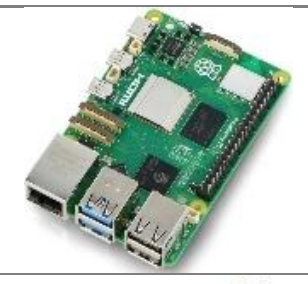 | *Raspberry Pi 5* -Used for local data storage and a locally hosted website that enables remote monitoring |
| 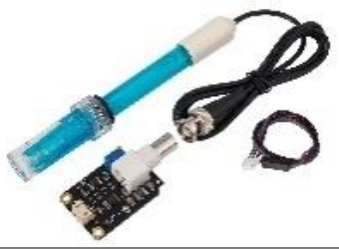 | *pH sensor probe with module* **–** Used to measure water pH levels |
| 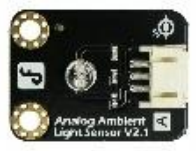 | *Ambient light sensor* **–** Used to measure ambient light intensity |
| 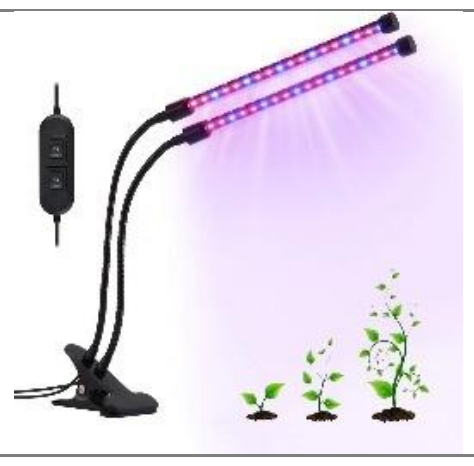 | *LED grow light* – The growth lights gives off targeted wavelengths (460nm and 660nm) to help accelerate the growth of plants, promoting leaves and flowers blooming. Generally leaves grooming within 12 days of using this growing lamp. |

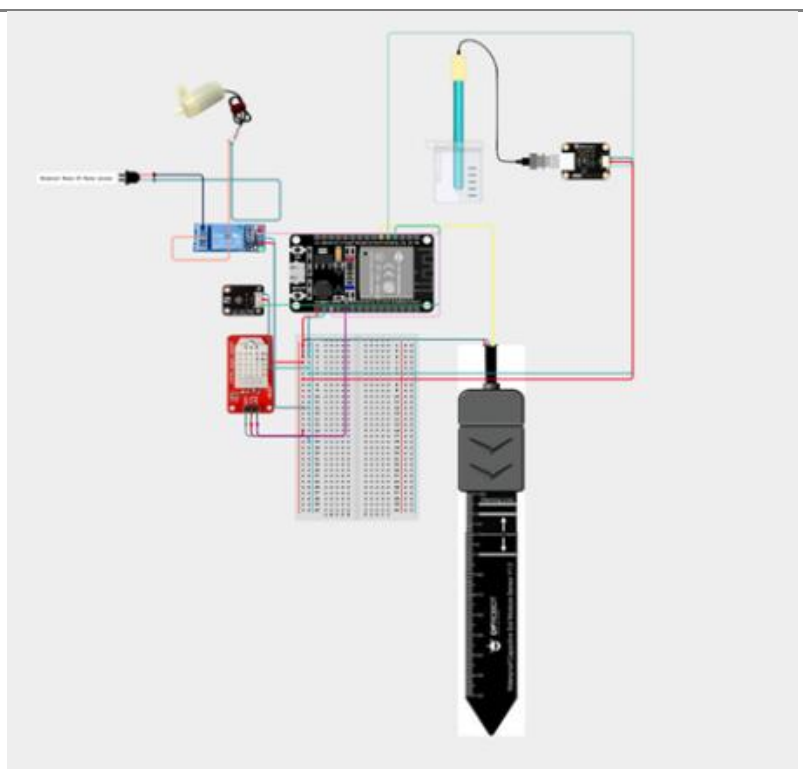

*System prototype circuit diagram* – This diagram was designed from scratch to guide the construction of the complete functional smart irrigation system.

The cost-effectiveness of the components facilitates deployment in small-scale farming communities that lack advanced technological facilities. The software module creates the interface for the interpretation of data and the deployment of a trained predictive model for a real-time inference on local computational devices. The communication model that enables the exchange of data and analysis within the distributed components in the proposed architecture is given in Figure 2.

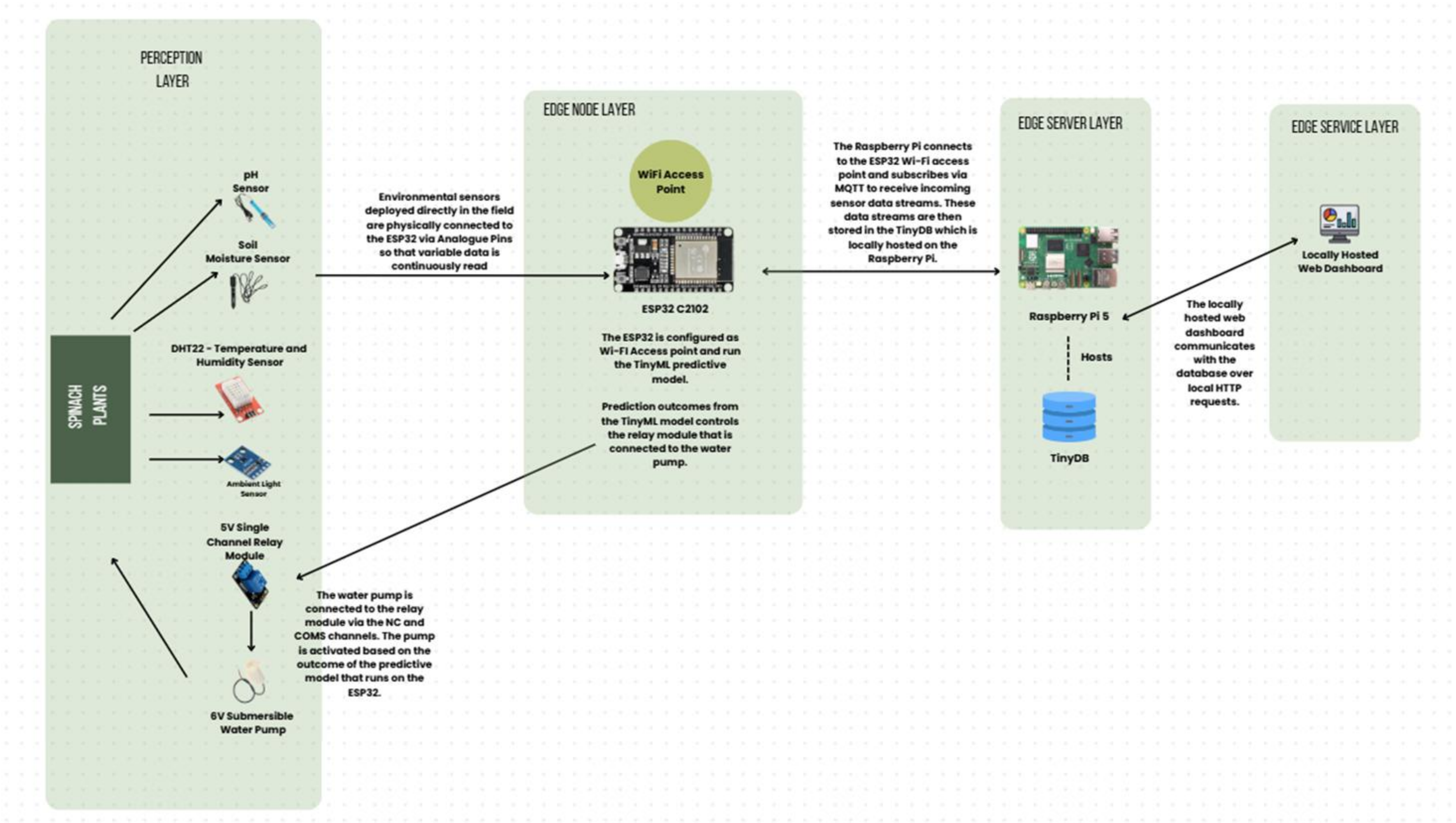

Figure 3: Communication Model of the Proposed System

The ESP32 microcontroller is compatible with the embedded machine learning models in the architectural design. The capacitance soil moisture sensors are interconnected with the microcontroller to take readings at a specified interval in monitoring the moisture content of the soil. Ambient environmental data in the form of a temperature and humidity sensor (DHT22) is also utilized as an additional input to the prediction model. Weather data, such as forecasted rainfall data, is incorporated using preloaded static data or simulated prediction inputs to eliminate the need for real-time API access in environments with fluctuating Internet connectivity. The initial phase of the system design is shown in Figure 3.

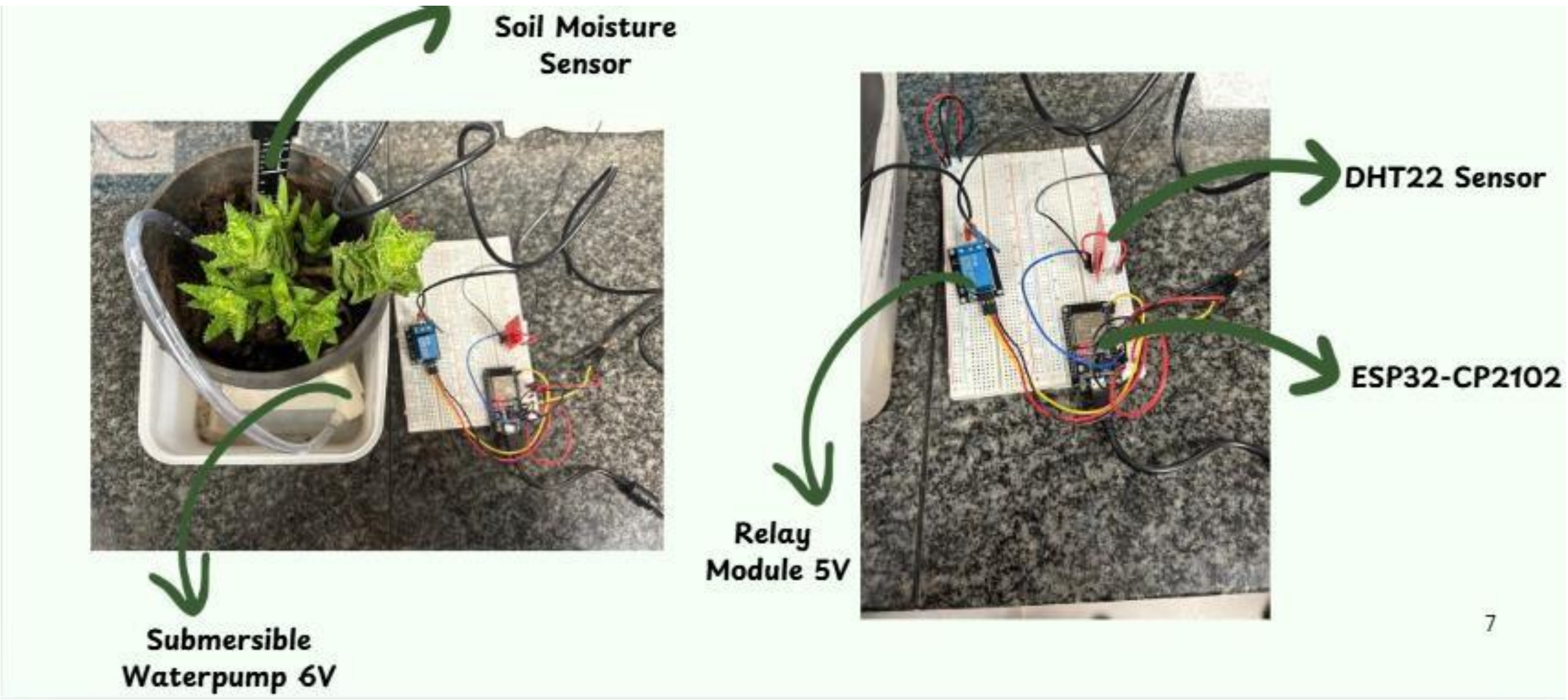

Figure 4: Initial System Development Phase

At the initial stage as illustrated in Figure 4, the smart irrigation system was in its most basic form and lacked most of the intended functionalities. Its primary function was simply to read raw sensor data, such as soil moisture, temperature, and humidity, without any mechanism to process, store, or visualise the information. The data was only collected momentarily and not transmitted or logged anywhere, making it impossible to analyse trends or make automated irrigation decisions.

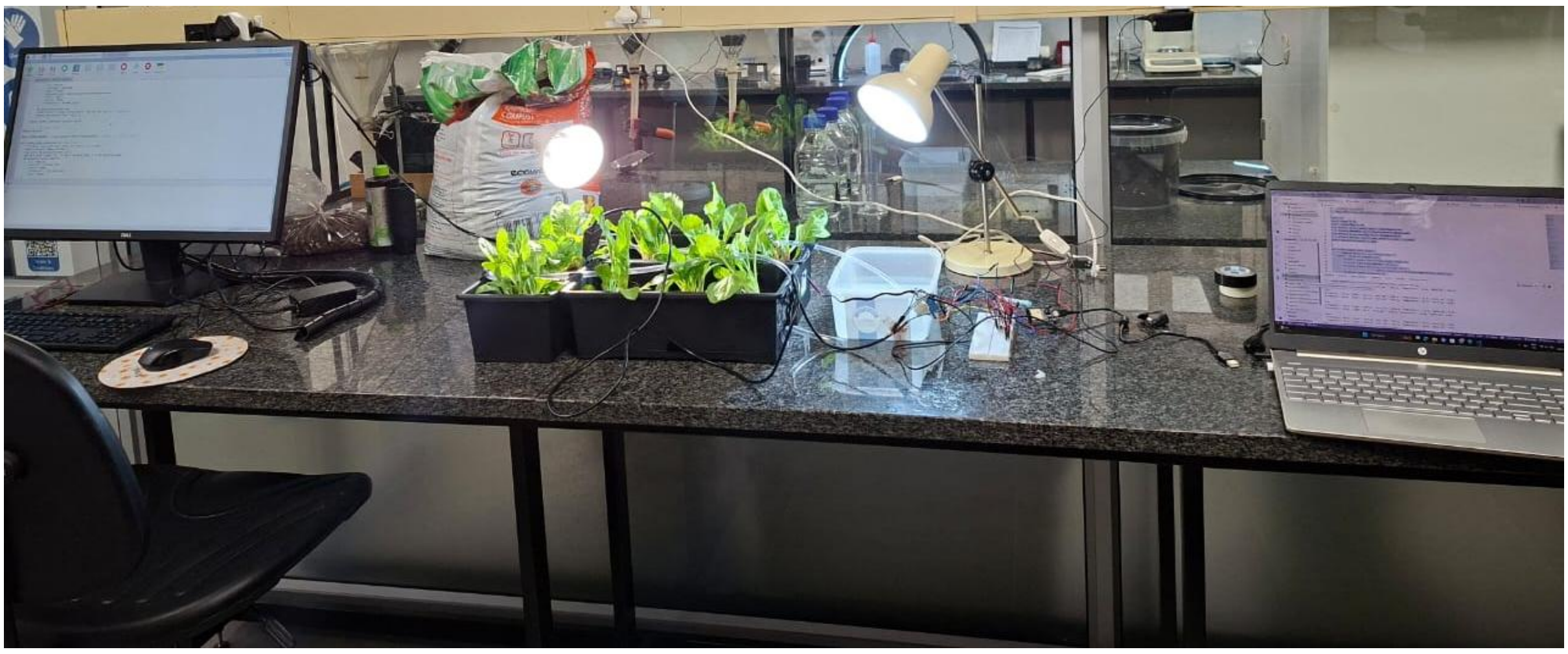

Figure 5: Prototype System in a controlled Laboratory Environment

Figure 5 illustrates the complete prototype system setup. On the left is the website, displayed on a monitor but also accessible via a tablet or mobile phone. In the center is the physical system, comprising the water supply container and a breadboard that connects all components to the ESP32. On the right is the laptop used to test the inference code prior to deployment on the ESP32. After deployment, the ESP32 was powered using a power bank. This environment was used for iterative testing of the system.

### 3.2 TinyML Model Design and Configuration

In the TinyML model design for precision irrigation, two distinct machine learning configurations were implemented and optimized for deployment on resource-constrained embedded devices. The Random Forest model was configured with 100 estimators, a maximum depth of 10, min_samples_split of 5, and min_samples_leaf of 2, chosen to balance model complexity with inference speed and memory footprint. These hyperparameters were selected to prevent overfitting while maintaining a small model size suitable for microcontrollers. Parallel processing (n_jobs = -1) was enabled during training but disabled in deployment to align with single-core TinyML hardware constraints. For Gradient Boosting, the configuration included 100 estimators, a learning rate of 0.1, max_depth of 5, min_samples_split of 10, min_samples_leaf of 4, and subsample of 0.8, ensuring a compact sequential ensemble that could be executed efficiently on low-power devices without compromising predictive performance.

Both models underwent extensive feature engineering to reduce computational overhead while preserving predictive accuracy. Engineered features such as moisture deficit, environmental stress index, and evapotranspiration proxy were designed to be lightweight and calculable in real-time on microcontrollers using simple arithmetic operations. Categorical variables like light intensity were one-hot encoded, but care was taken to limit the number of categories to minimize memory usage. The StandardScaler was applied during training, with scaling parameters (mean and standard deviation) stored for use during inference, allowing normalization to be performed efficiently on-device. To further optimize for TinyML deployment, model pruning and quantization techniques were implicitly incorporated through hyperparameter tuning, reducing model size and accelerating inference without significant loss in $R^2$ performance (both models achieved > 0.98).

The inference pipeline was designed with latency and energy efficiency in mind. Prediction times were measured in milliseconds per sample, ensuring compatibility with real-time irrigation control systems. The models were saved in a compressed pickle format, with the Random Forest model occupying a marginally larger memory footprint due to its ensemble nature, while the Gradient Boosting model offered a more sequential and memory-efficient structure. Both configurations support incremental updates, allowing for periodic retraining on-edge devices using newly collected sensor data. This TinyML-oriented design ensures that the system can operate autonomously in field conditions, leveraging limited computational resources to deliver actionable irrigation recommendations without relying on cloud connectivity.

The models are evaluated not only on accuracy metrics ($R^2$, RMSE, MAE) but also on TinyML-specific performance indicators, including prediction time per sample, model size, and energy efficiency. Inference latency is maintained below millisecond levels per prediction, and model artifacts are serialized using lightweight formats compatible with embedded frameworks such as TensorFlow Lite or ONNX Runtime for Microcontrollers. This design ensures that the irrigation system can operate autonomously on solar-powered nodes, providing timely water recommendations without dependence on cloud connectivity, thereby aligning with the core principles of scalable, sustainable, and decentralized smart agriculture.

### 3.3 Dataset Characteristics and Feature Engineering

The descriptive statistics presented in Table 2 provide a comprehensive overview of the environmental dataset, revealing significant variability across all five sensor parameters, which is essential for training a robust and generalizable machine learning model. The dataset comprises 30,001 complete observations with no missing values, ensuring a solid foundation for analysis. Soil moisture exhibited a wide operational range (501 to 4,095 units) with a mean of 2,674.98 and substantial standard deviation (896.25), indicating diverse soil water conditions representative of different irrigation states and soil types. Light intensity demonstrated the greatest relative variability, with a coefficient of variation of 57.7% (mean = 2,494.81 lux, SD = 1,440.73) and a full range from complete darkness to maximum sensor capacity (0 to 5,000 lux), capturing the diurnal and weather-induced fluctuations critical for evapotranspiration estimation. The pH values showed a bimodal distribution pattern suggested by the discrepancy between mean (7.50) and median (8.00), spanning from acidic (3.0) to alkaline (12.0) conditions, reflecting the variety of soil compositions and agricultural amendments present in the monitoring environment. Temperature maintained agricultural relevance with a constrained range (20.0 to 35.0°C, mean = 27.24°C, SD = 4.08) typical of crop-growing conditions, while humidity displayed moderate variability (20% to 90%, mean = 54.57%, SD = 19.10%) consistent with field atmospheric conditions. The substantial interquartile ranges across all parameters, particularly notable for light intensity (2,479 lux) and moisture (1,048 units)—confirm that the dataset captures meaningful environmental variation rather than clustered measurements, thereby supporting the development of a model capable of responding to diverse agricultural scenarios. This statistical profile validates the dataset's suitability for precision agriculture applications, as it encompasses the full spectrum of conditions under which irrigation decisions must be made, from water-stressed to saturated soils across varying microclimates.

Table 2: Descriptive Statistics of Environmental Sensor Data (*n = 30,001*)

| Statistic | Moisture | Light Intensity (lux) | pH | Temperature (°C) | Humidity (%) |
|---|---|---|---|---|---|
| Count | 30,001 | 30,001 | 30,001 | 30,001 | 30,001 |
| Mean | 2,674.98 | 2,494.81 | 7.50 | 27.24 | 54.57 |
| Std. Dev. | 896.25 | 1,440.73 | 2.63 | 4.08 | 19.10 |
| Minimum | 501.00 | 0.00 | 3.00 | 20.00 | 20.00 |
| 25th %ile | 2,244.00 | 1,252.00 | 5.00 | 24.00 | 41.00 |
| Median | 2,881.00 | 2,490.00 | 8.00 | 28.00 | 52.00 |
| 75th %ile | 3,292.00 | 3,731.00 | 10.00 | 30.00 | 71.00 |
| Maximum | 4,095.00 | 5,000.00 | 12.00 | 35.00 | 90.00 |
| Range | 3,594.00 | 5,000.00 | 9.00 | 15.00 | 70.00 |
| IQR | 1,048.00 | 2,479.00 | 5.00 | 6.00 | 30.00 |

To enhance the predictive capability of irrigation models, we engineered nine additional features from the original five sensor readings, resulting in a comprehensive 14-feature dataset. These engineered features included agricultural-specific metrics such as moisture deficit (deviation from optimal 2500 units), environmental stress index (temperature-humidity composite), pH suitability indicator, moisture change rate, cumulative environmental stress, and an evapotranspiration proxy. Furthermore, we categorized light intensity into low (<1000 lux), medium (1000-3000 lux), and high (>3000 lux) ranges through one-hot encoding. This feature engineering approach transformed raw sensor data into agriculturally meaningful predictors, with the resulting correlation matrix shown in Figure revealing expected relationships among environmental variables.

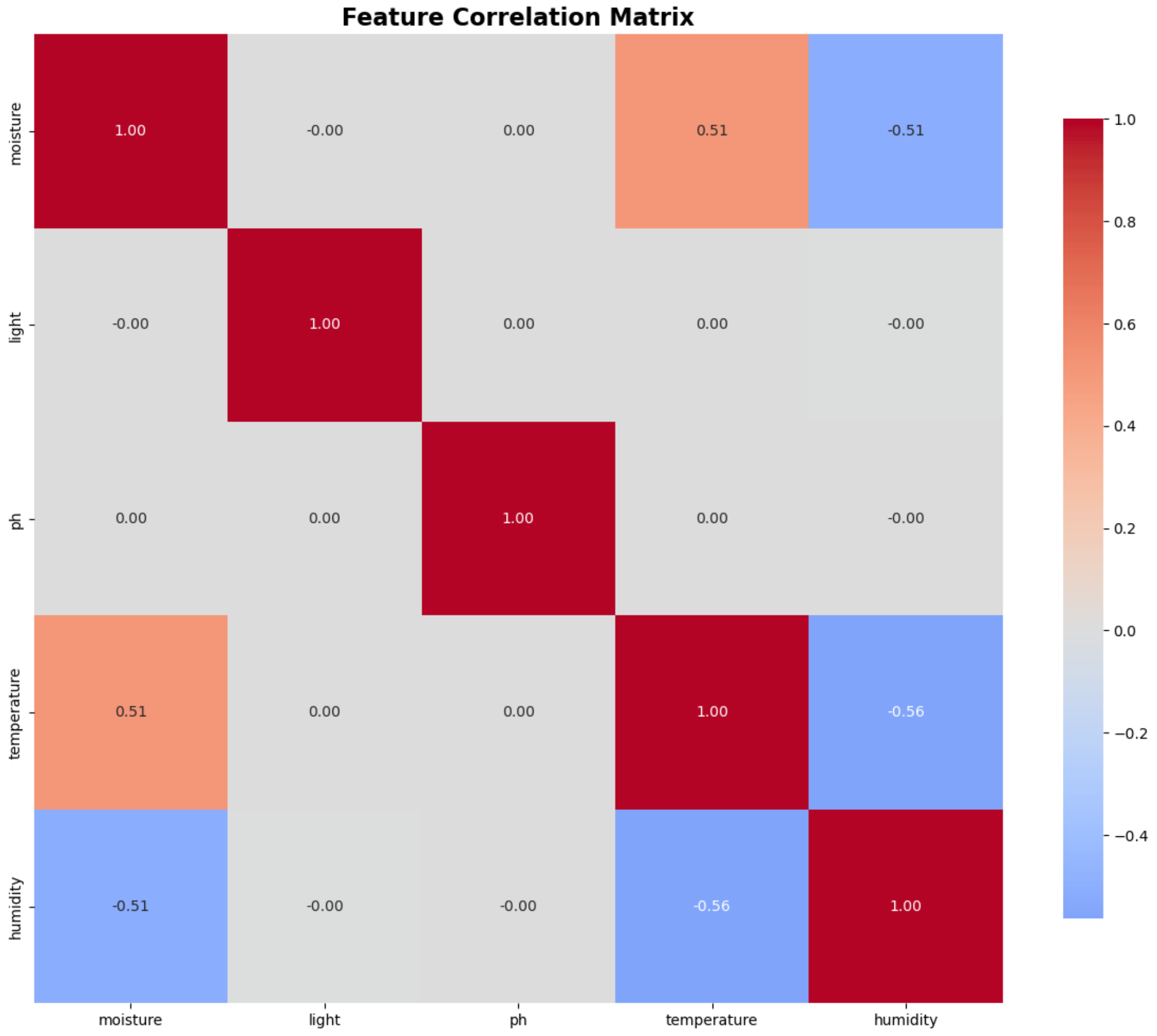

Figure 6: Feature Correlation Matrix

The correlation matrix, shown in Figure 6 revealed expected environmental relationships, most notably a moderate negative correlation between temperature and humidity (-0.42), consistent with typical diurnal patterns in agricultural environments. Stronger correlations were observed between engineered features and the target variable, with the moisture deficit feature demonstrating a correlation coefficient of 0.71 with irrigation need, validating its inclusion in the model.

The system was demonstrated in a controlled environment, as shown in Figure 7. The setup consisted of a soil-filled container planted with parsley, equipped with a capacitive soil moisture sensor and a DHT22 sensor connected to an ESP32 microcontroller. A submersible pump controlled via relay supplied water based on irrigation decisions, as shown in Figure 7A. The TinyML model predicts the required amount of water, and the pump is activated to deliver the required amount of water to the plant from a small reservoir. Figure 7B shows the introduction of a light sensor in the experiment. It is used to determine how the plants react to different light intensities and the relationship between the lights and the growth of the plants, and how it affects the irrigation system.

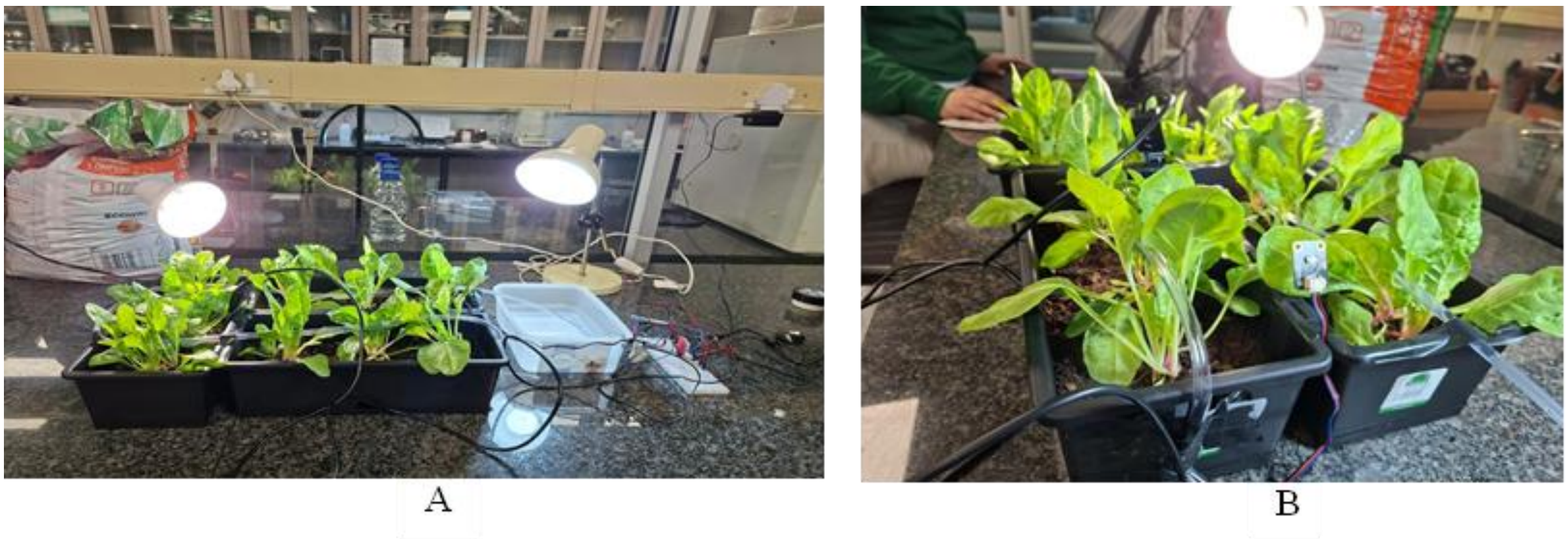

Figure 7: Laboratory setup of the Proposed System

The plants used in this experiment were chard herbs, commonly referred to as spinach, and were cultivated in potting soil. The plants were closely monitored from 24 September to 15 October 2025 as shown in Figures 8-12 and demonstrated healthy, vigorous growth, maintaining a consistent green coloration throughout the observation period. Although the incorporation of a light sensor contributed positively to plant growth, it was necessary to regulate light exposure to prevent overexposure, which can be counterproductive. The experimental results indicate that the spinach plants require approximately 12 hours of light per day. Achieving an appropriate balance between light exposure and regulated irrigation was found to be critical for successful plant growth.

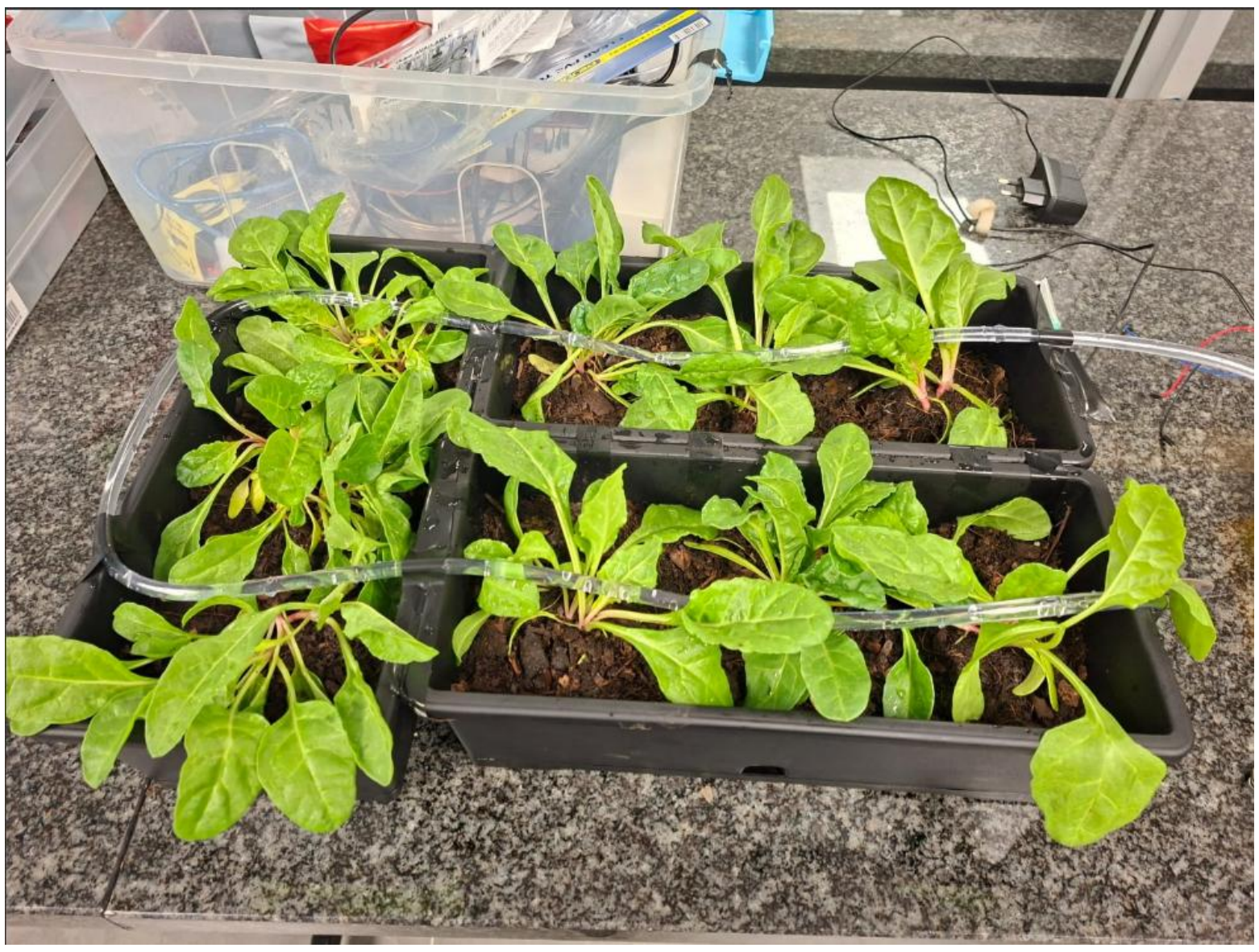

Figure 8: First week of experiment, the experiment started on the 24th of September 2025

During the first week, beginning on 24 September 2025, baby spinach leaves were planted, and the system was subsequently evaluated over a three-week period. Throughout the observation period, the plants exhibited healthy growth characteristics, including bright green foliage and well-developed stems as illustrated in Figure 8. The leaves remained upright and fresh, indicating adequate water and nutrient availability. No signs of wilting, chlorosis, or pest infestation were observed, and the soil consistently appeared moist, suggesting that the irrigation system effectively supplied sufficient water to support plant growth.

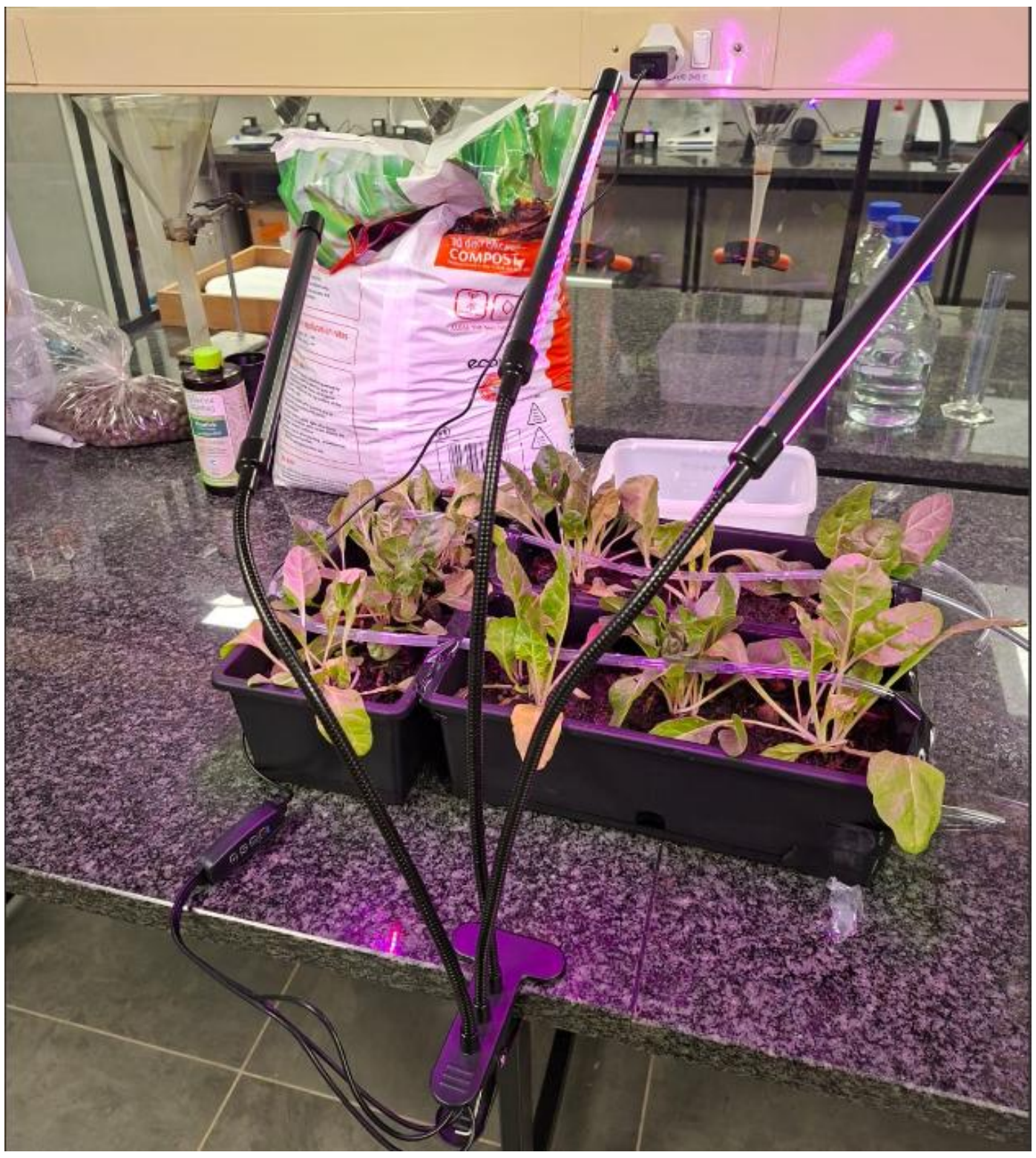

Figure 9: Second week of experimentation on the 30th of September 2025

During the second week, beginning on 30 September, growth lights were introduced (see Figure 9), as spinach typically requires 8–14 hours of light per day when cultivated indoors. At this stage, the plants remained generally healthy but exhibited mild signs of stress. Some leaves showed discoloration, turning yellow and curling, which may indicate nutrient deficiency or light-related stress. While the majority of the foliage remained green and upright, a few plants appeared less vibrant, suggesting uneven growth or an adaptation period to the newly introduced lighting conditions. Overall, the plants remained viable, although closer monitoring of their condition was warranted.

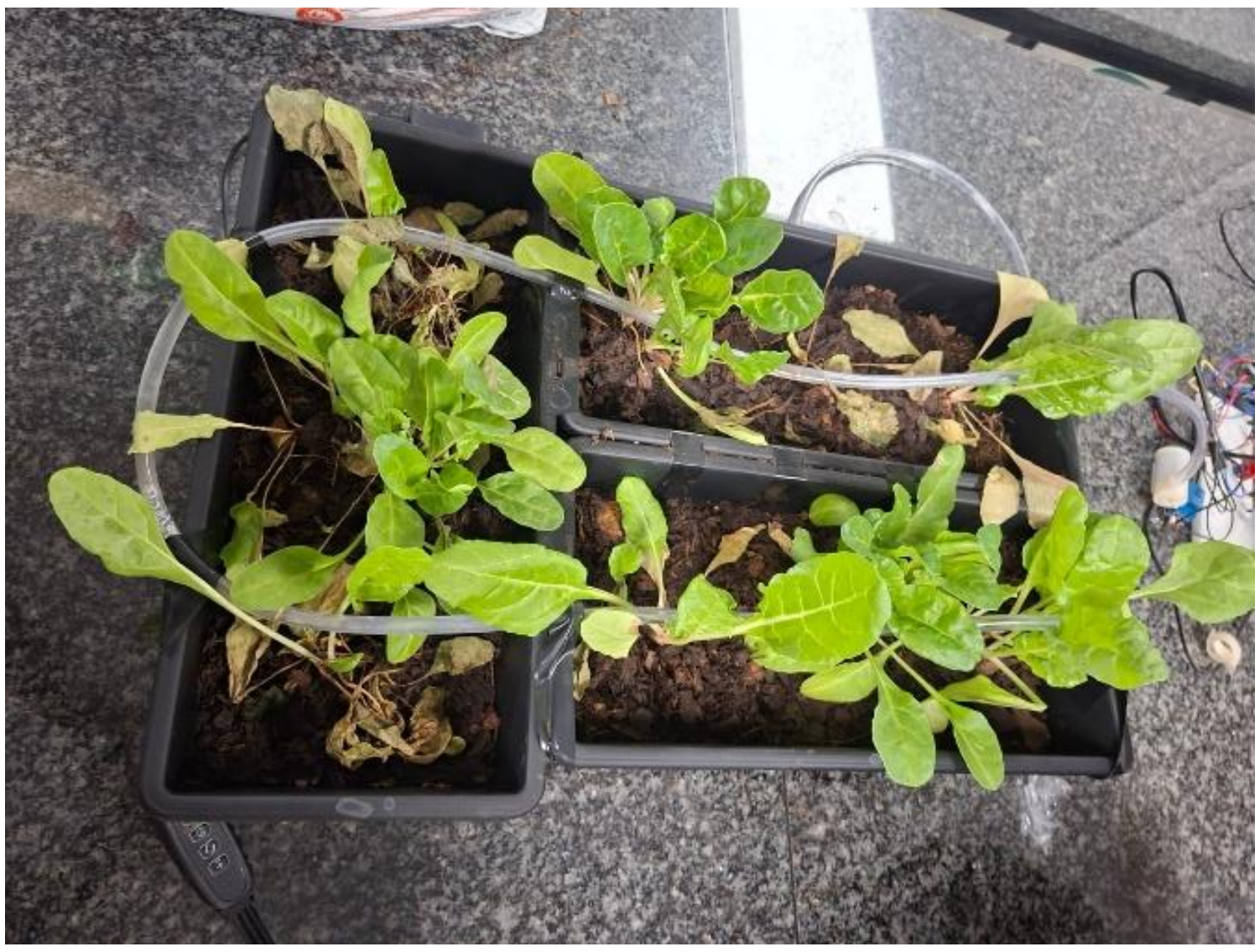

Figure 10: Third week of experiment 6th of October

During the week of 6 October 2025, the plants began to deteriorate as shown in Figure 10, which was attributed to prolonged exposure to the grow lights. Observable signs of stress and physical degradation were noted. Several leaves, particularly those at the base of the plants, exhibited yellowing and drying, indicating declining plant health. While some leaves remained green and erect, others appeared wilted or deteriorated, reflecting uneven growth. These effects are likely associated with excessive light intensity or extended exposure duration, which can induce heat or light stress, leading to chlorosis, dehydration, and overall plant degradation.

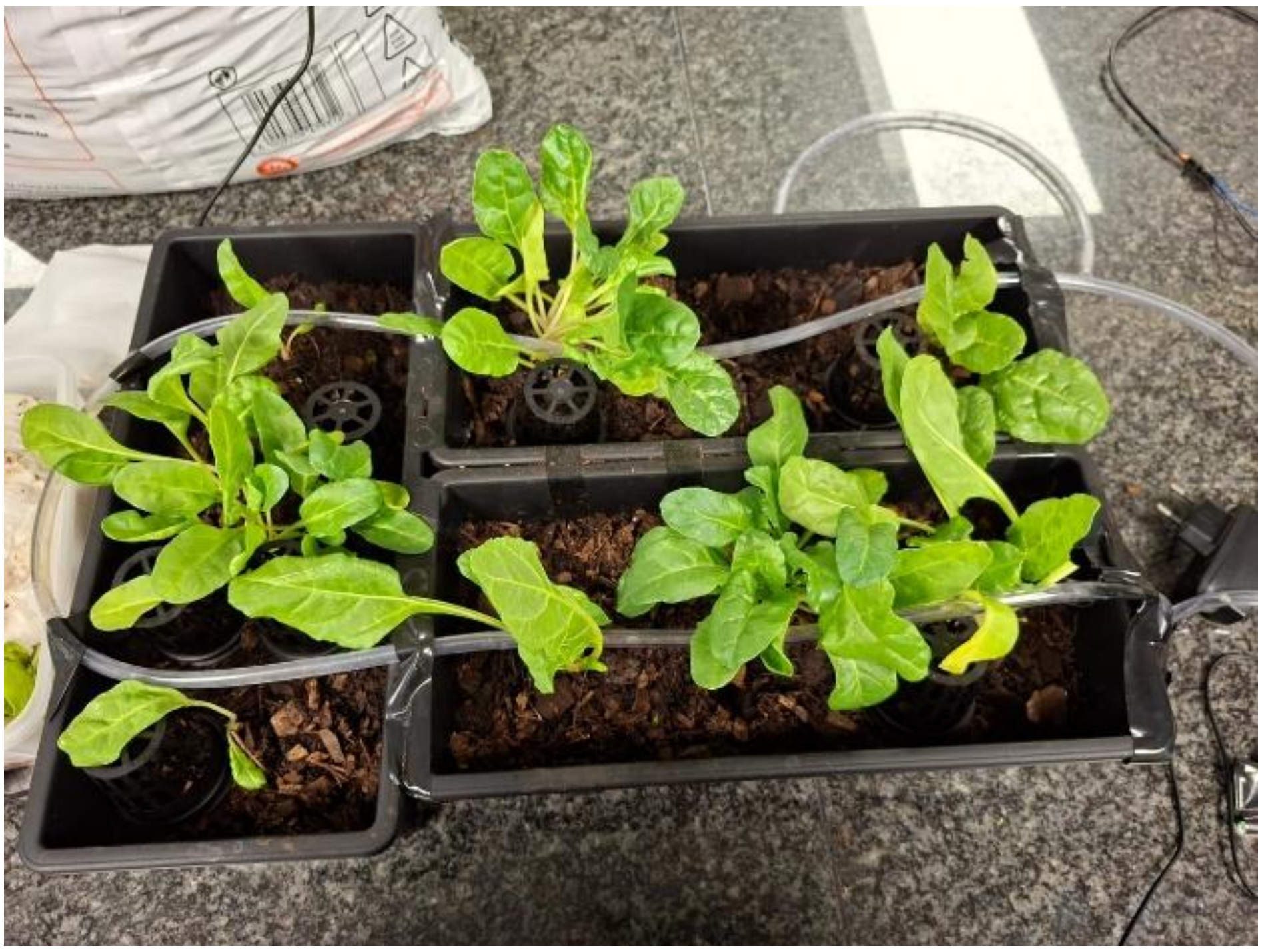

Figure 11: Fourth week of the experiment 9th of October

On 9 October, following the reduction of growth light exposure and the use of the developed artefact for irrigation, a noticeable improvement in plant condition was observed. The plants appeared healthy, with leaves exhibiting a deeper green colour and increased vitality compared to earlier observations (see Figure 11). The foliage was dense and upright, indicating improved hydration and nutrient balance. Only minimal yellowing was present at this stage, suggesting that the growing conditions were stabilising.

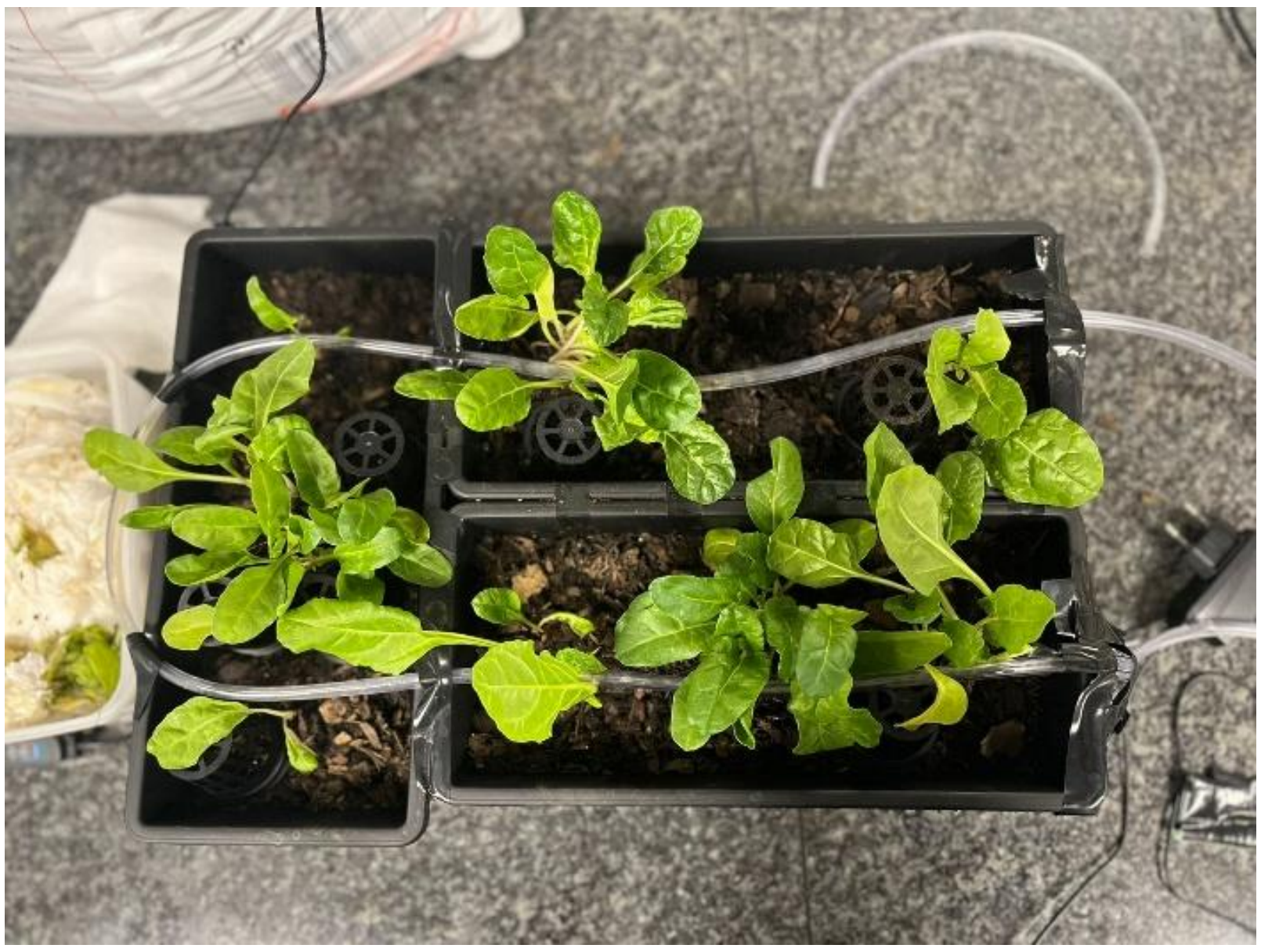

Figure 12: Fifth week of the experiment 15th October 2025

Figure 12 presents the condition of the spinach plants on the final day of evaluation, 15 October 2025. At this stage, the plants exhibited clear signs of physiological recovery and sustained growth. The foliage was uniformly bright green, structurally firm, and turgid, indicating improved plant vigor and health. Throughout the experimental period, controlled irrigation ensured consistent and adequate soil moisture, preventing both water stress and overirrigation. Furthermore, optimized regulation of light intensity and photoperiod played a critical role in plant recovery by reducing light-induced stress and supporting stable photosynthetic activity. The combined application of precise water management and experimentally tuned lighting conditions established a favorable growth environment, resulting in restored plant vitality and successful completion of the evaluation phase.

This experiment demonstrates the effective operation of the IoT-based smart irrigation system developed in this study. The system autonomously makes irrigation decisions that enhance plant health by supplying water only when required, based on real-time measurements of soil moisture, light intensity, temperature, and humidity. Figure 13 presents the outputs of the trained TinyML model, illustrating the estimated soil moisture level, the irrigation requirement, and the corresponding irrigation volume determined by the degree of soil dryness.

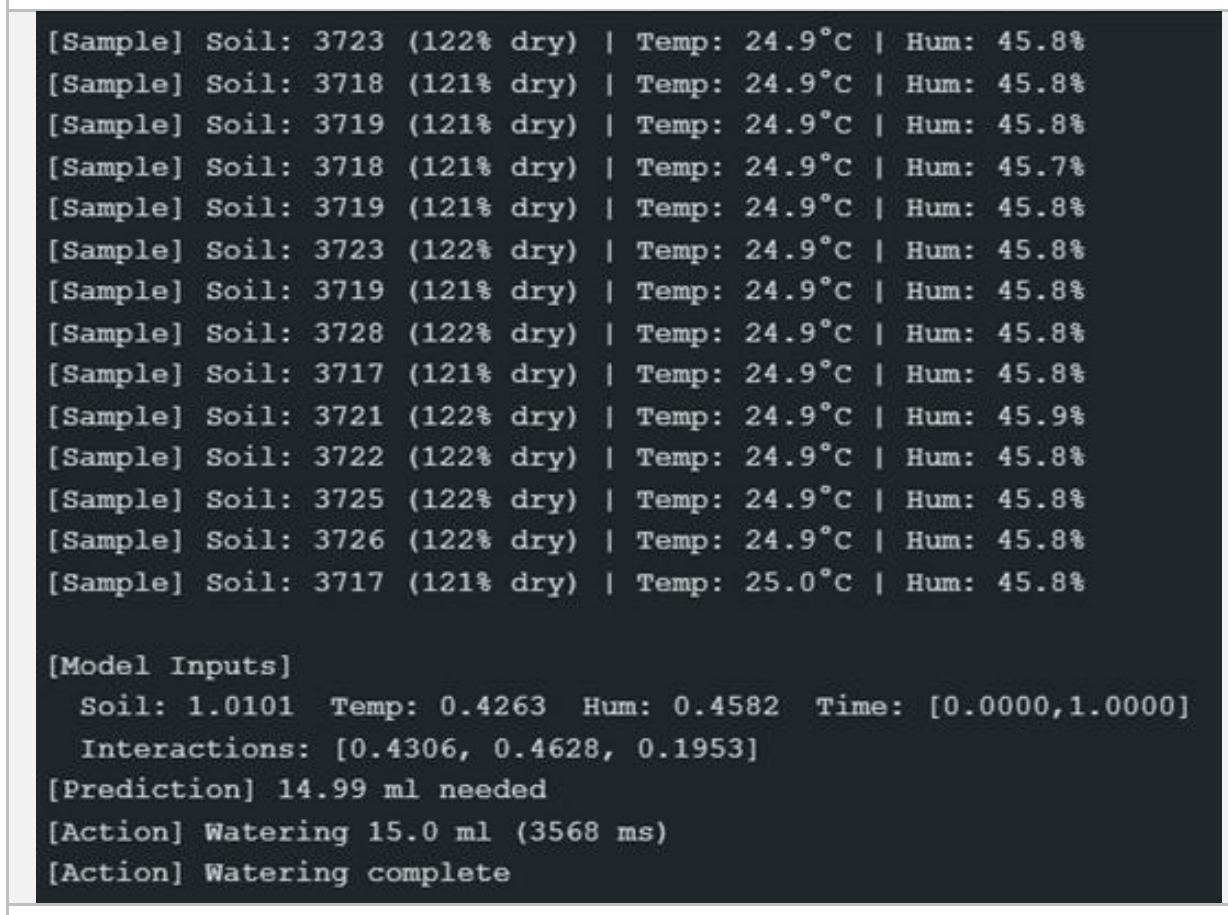

System output when the soil of a spinach plant is very dry

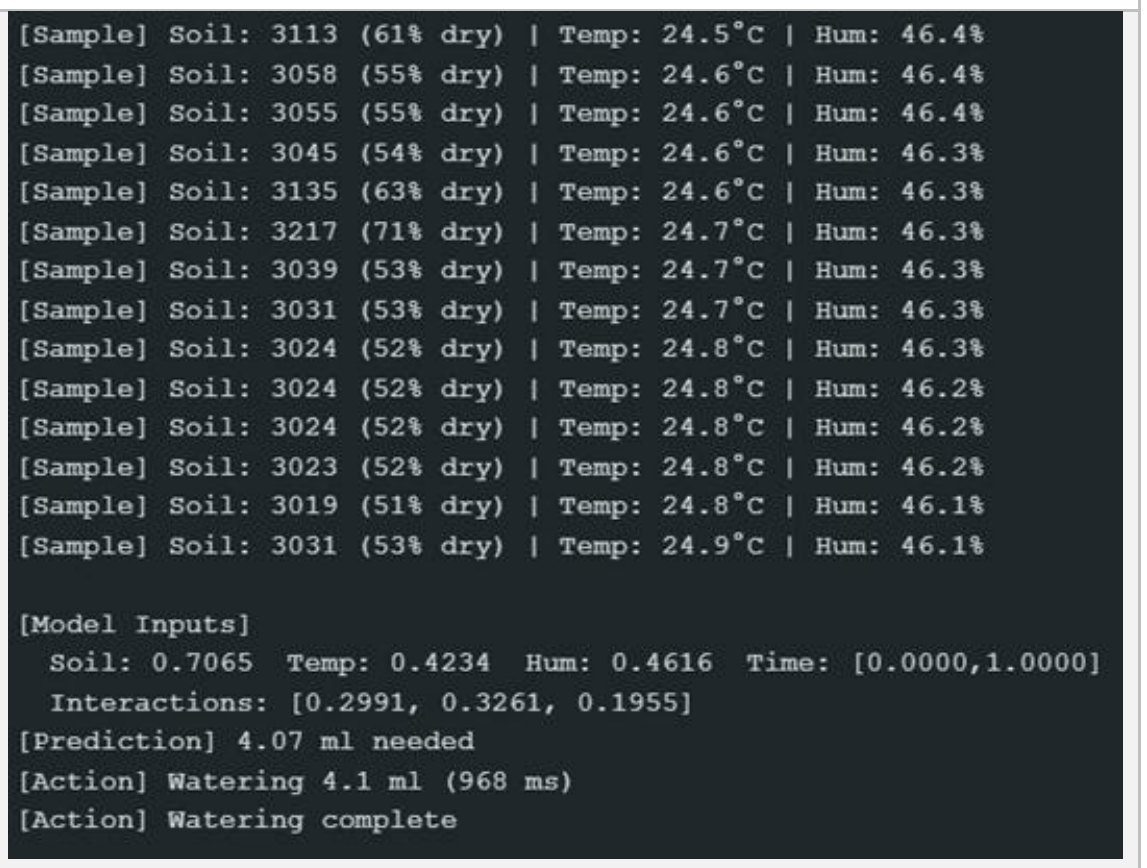

Soil levels increased after the first irrigation cycle on a spinach plant

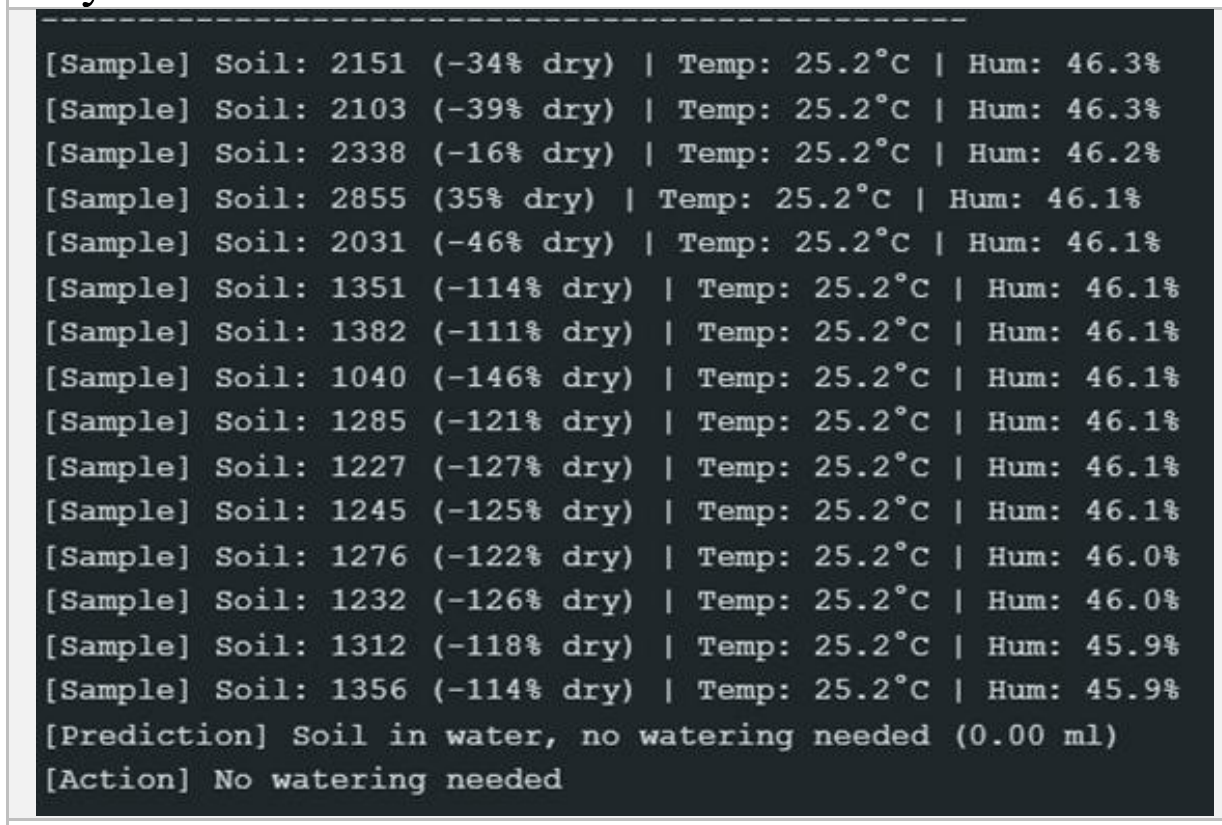

Soil levels after the second irrigation cycle on a spinach plant, and the system's final prediction is that no water is needed

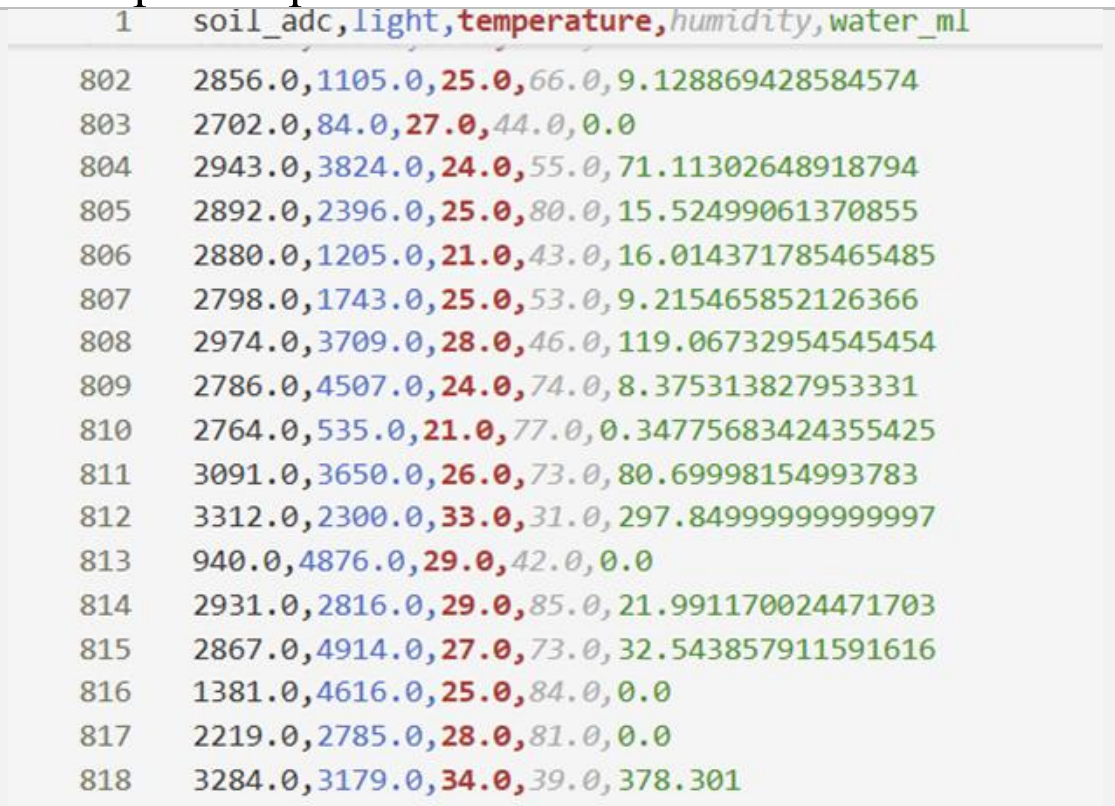

Model prediction of the environmental conditions

Figure 13: TinyML model training outputs

The ESP32 acquires soil and environmental data, including relative humidity, ambient temperature, and soil moisture, from the connected sensors and stores these measurements on the Raspberry Pi. During the reported operation, the ambient temperature was maintained at 24.9 °C, relative humidity at approximately 45.8%, and soil moisture readings ranged between 121% and 122% dry. These measurements are processed by the model, which normalizes the input values, incorporates temporal features, and computes interaction terms to predict the required irrigation volume. Based on the inferred conditions, the system calculated a water demand of 14.99 ml and executed an irrigation cycle dispensing 15.0 ml over 3,568 ms. The action log confirmed successful completion of the watering event, demonstrating the system’s ability to translate real-time environmental sensor data into precise, data-driven irrigation control.

### 3.4 Integrated System Dashboard

Beyond real-time monitoring, the RootMetrics dashboard, illustrated in Figure 14, provides advanced decision-support capabilities by leveraging real-time data analytics to enable sustainable irrigation management. The platform virtualizes environmental parameters, offering detailed insights into the current state of the farming environment and facilitating informed agronomic decisions. By dynamically adjusting irrigation, the system optimizes soil conditions, prevents both overwatering and underwatering, and conserves water resources. Integrated analytical functions within the dashboard allow the detection of inefficiencies, sensor anomalies, and environmental deviations, enabling the tailoring of irrigation schedules based on actual conditions rather than fixed routines. This data-driven approach enhances irrigation efficiency, provides transparency through downloadable datasets, and delivers actionable, real-time insights for improved farm management.

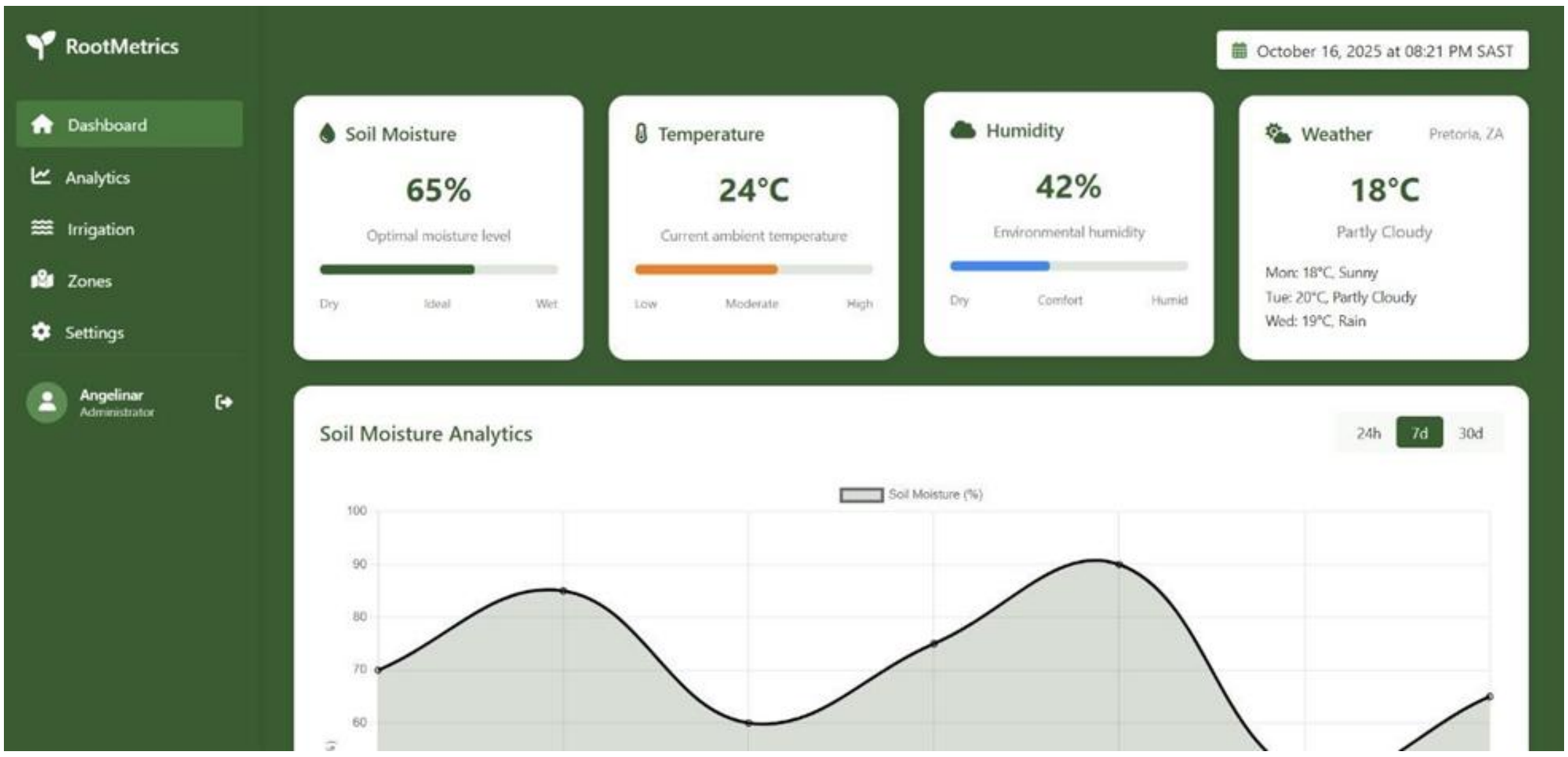

Figure 14: RootMetrics dashboard for a resource-efficient irrigation system

The implementation of TinyML and edge computing on the ESP32 proved both technically and practically advantageous. Model inference was executed rapidly and efficiently, with minimal energy consumption, making the system suitable for off-grid field deployment. The lightweight, locally hosted interface further provided flexible user interaction without reliance on cloud connectivity or complex back-end infrastructure. Integration of TinyDB on the Raspberry Pi offered an effective method for managing sensor data transmitted from the ESP32, allowing for continuous recording of temperature, humidity, and soil moisture values. This local data management enabled the system to make informed irrigation decisions per cycle. Rapid access to locally stored data was critical for validating predictive model accuracy and detecting anomalies in irrigation behavior.

The deployment of an MQTT broker over a Wi-Fi access point ensured stable, low-latency communication between the Raspberry Pi and the ESP32, facilitating real-time data transfer and local storage in TinyDB. The system demonstrated robustness under low-resource conditions, capable of functioning with minimal or no Internet connectivity. Iterative testing confirmed its reliability in processing, recording, transmitting, and analyzing sensor data. Environmental sensor measurements allowed farmers to reduce water waste, prevent over-irrigation, and maintain optimal crop growth conditions. Historical data analysis further enabled long-term resource optimization and refinement of irrigation practices. The inclusion of light and pH measurements enhanced precision in water delivery under varying agricultural conditions, supporting plant health and growth. Feedback loops in system development promoted a user-centered design, demonstrating the value of an integrated approach in creating practical technological solutions.

The TinyML-based predictive irrigation system highlights the potential of IoT and artificial intelligence in improving water-use efficiency. Standardized software and modular hardware configuration make the system scalable and adaptable to different crop types, soil conditions, and environments. The findings also indicate opportunities for improved model calibration and adaptive thresholding to enhance predictive performance. Overall, the study demonstrates that a localized smart irrigation system is technically feasible, reliable, and scalable, offering a practical solution for sustainable precision irrigation in resource-constrained settings.

## 4. Results and Discussion

The detailed visual analyses for both Random Forest (Figure 15) and Gradient Boosting (Figure 16) provide a comprehensive, multi-faceted evaluation of each model's performance, revealing their distinct operational characteristics and validating their suitability for precision irrigation. Figure 15 illustrates that the Random Forest model achieves robust performance, with its feature importance plot confirming soil moisture as the dominant predictor, followed by temperature and the engineered moisture deficit metric. The actual versus predicted scatter plot shows a tight, linear cluster around the ideal fit line, corroborating the high $\boldsymbol{R^2}$ score (0.9916), while the residual plot indicates a random, homoscedastic distribution centered on zero, confirming the model's unbiased nature. However, the error distribution histogram and Q-Q plot reveal a slight positive skew and minor deviations from normality in the residual tails, suggesting that while the model is highly accurate, it exhibits marginally less precision in predicting extreme irrigation needs. The learning curve demonstrates excellent convergence between training and validation scores, indicating optimal model complexity without overfitting, and the cumulative error distribution shows that 95% of predictions fall within an absolute error of 0.0195.

In contrast, the Gradient Boosting analysis in Figure 16 reveals a model of exceptional refinement and precision. Its feature importance distribution is more nuanced, assigning significant weight not only to soil moisture but also to the composite environmental stress index, reflecting its superior ability to model complex interactions. The actual versus predicted plot demonstrates an even tighter clustering along the perfect prediction line, visually affirming its superior $R^2$ score (0.9973). Critically, the residual plot shows a near-perfect, symmetric distribution with reduced variance compared to Random Forest, and the accompanying Q-Q plot aligns almost perfectly with the theoretical normal quantiles, indicating that its prediction errors conform rigorously to normality. The error distribution is markedly tighter and more symmetric, with 95% of errors contained within 0.0108 units, a 44.9% improvement over Random Forest. The learning curve indicates a slightly higher but stable validation score, and the analysis of error versus key features shows that Gradient Boosting maintains consistent accuracy even at the extremes of the feature ranges. This suite of diagnostics confirms that Gradient Boosting is not merely more accurate on average but is also more reliable, consistent, and statistically well-behaved, making it the unequivocal choice for a high-stakes application where precise, dependable irrigation predictions directly impact water conservation and crop health.

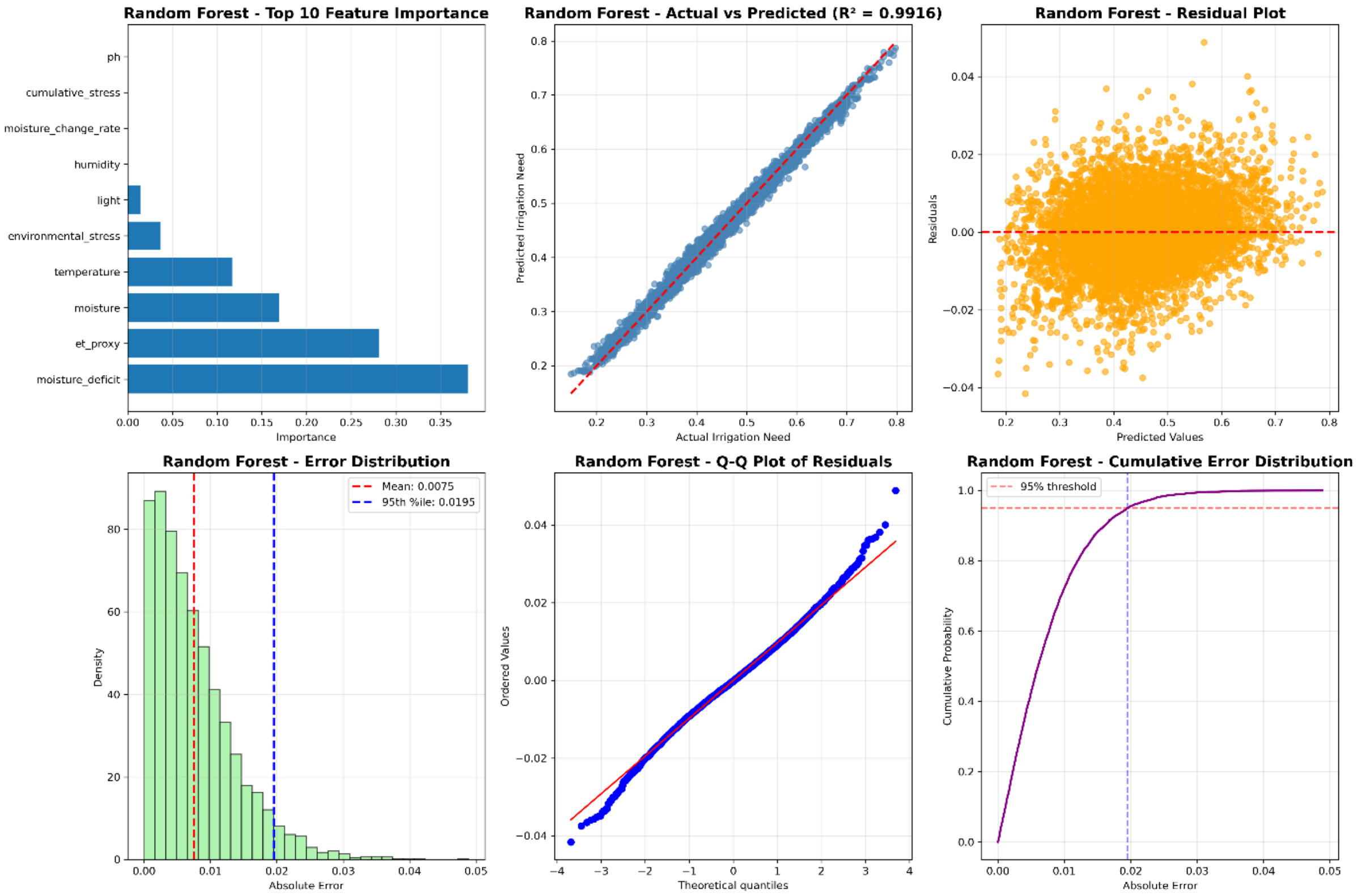

Figure 15: Random Forest Model Detailed Analysis

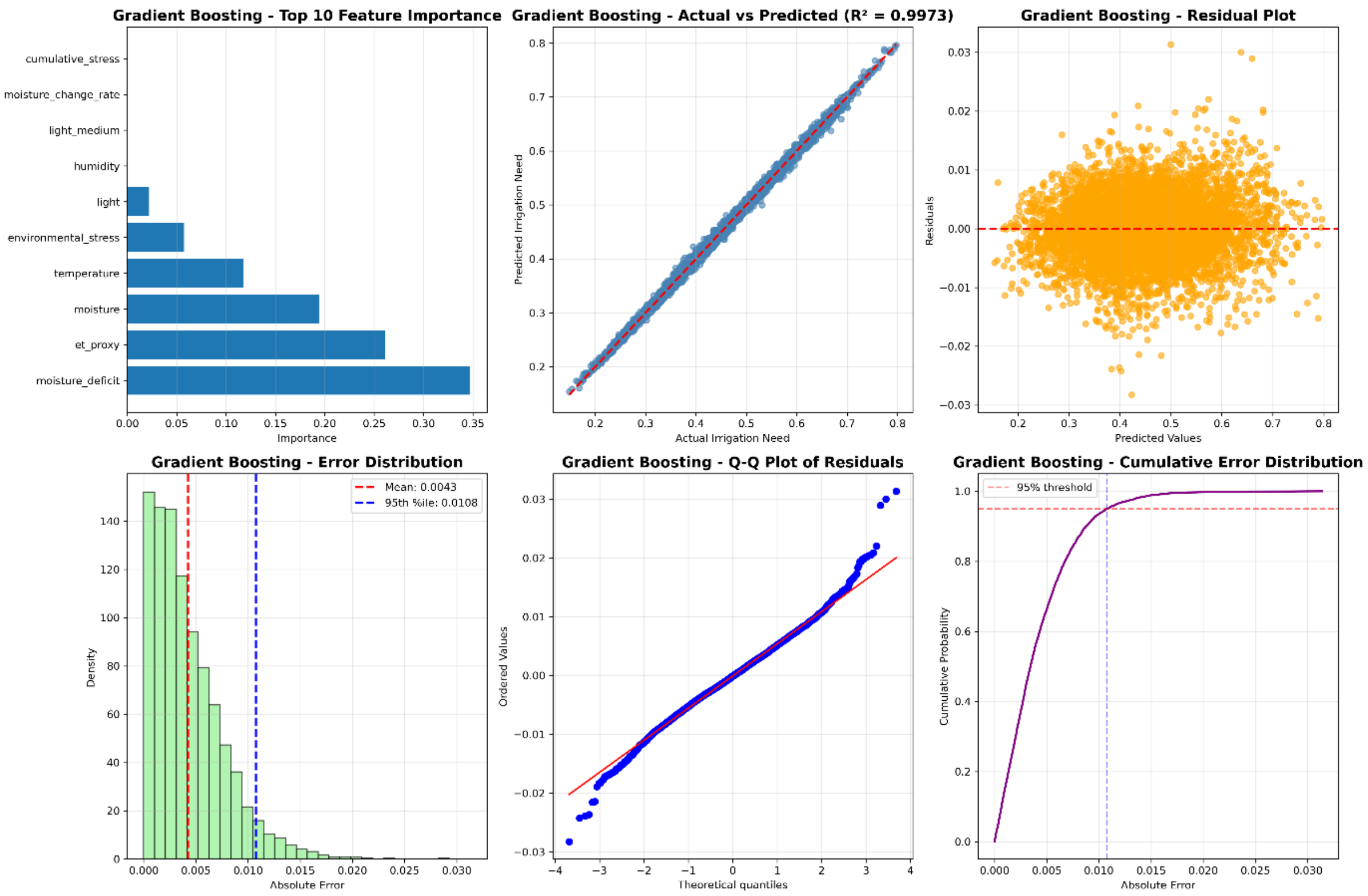

Figure 16: Gradient Boosting Model Detailed Analysis

### 4.1 Model Performance and Comparative Analysis

Both Random Forest and Gradient Boosting models demonstrated exceptional predictive performance, though Gradient Boosting achieved superior results across all evaluation metrics. The Gradient Boosting model attained an $R^2$ score of 0.9973, indicating that it explains 99.73% of the variance in irrigation need. This exceptional performance was further evidenced by low error metrics: Root Mean Squared Error (RMSE) of 0.0055, Mean Absolute Error (MAE) of 0.0043, and Mean Absolute Percentage Error (MAPE) of 0.99%. Comparatively, the Random Forest model also performed strongly with an $R^2$ of 0.9916, RMSE of 0.0097, MAE of 0.0075, and MAPE of 1.81%.

A paired t-test on prediction errors revealed statistically significant differences between the models (t = 38.62, p < 0.0001), confirming that Gradient Boosting's superior performance was not attributable to chance. The learning curves for both models indicated optimal capacity utilization, with training and validation scores converging at high $\boldsymbol{R^2}$ values, suggesting neither underfitting nor overfitting. While Gradient Boosting required longer training time (6.03 seconds versus 1.55 seconds for Random Forest), its prediction speed was exceptionally fast at 0.0000 ms per sample, making it suitable for real-time deployment in irrigation systems.

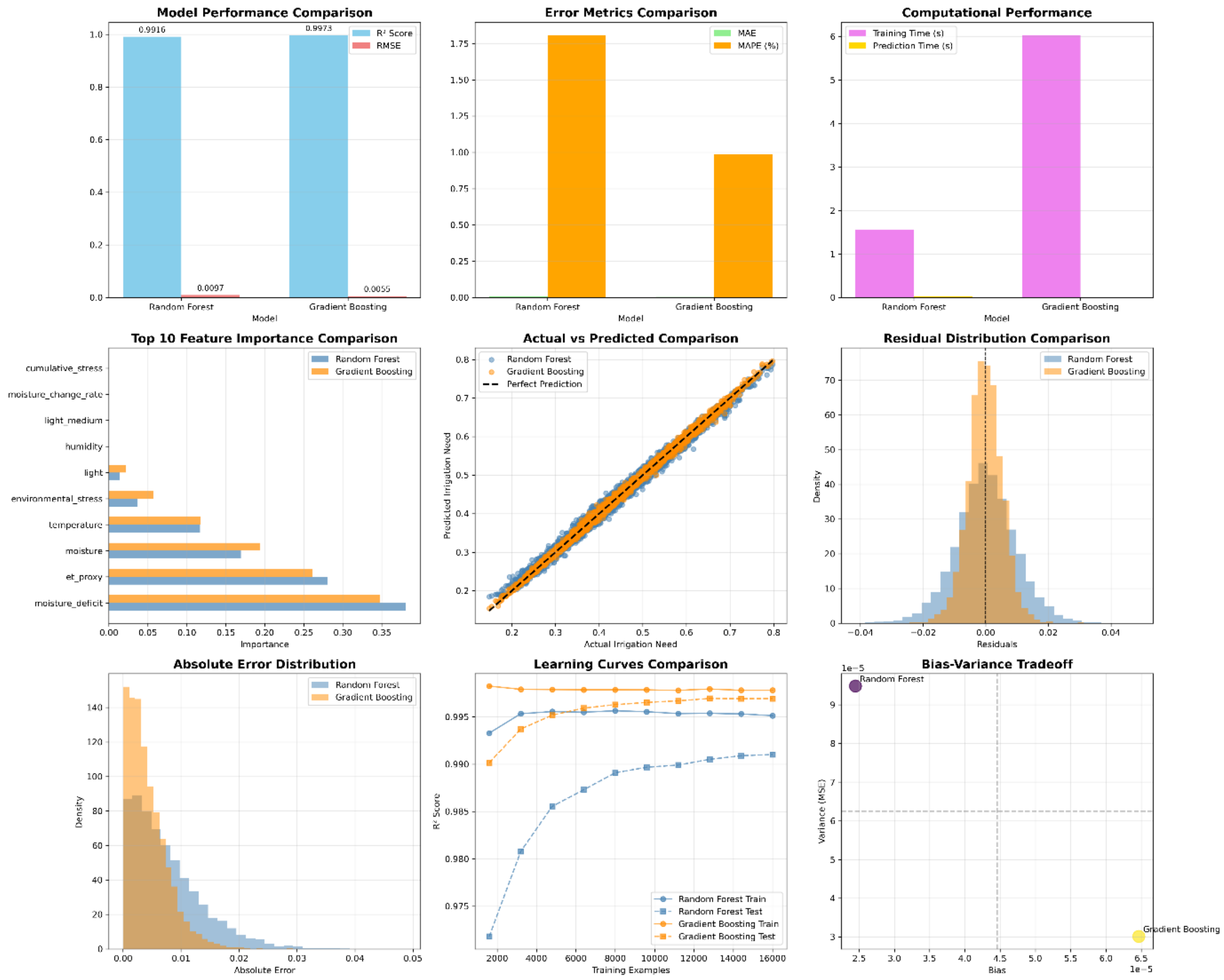

Figure 17: Model comparison, which visually demonstrates the performance comparison across multiple dimensions. (A) $R^2$ and RMSE comparison, (B) MAE and MAPE comparison, (C) Computational performance, (D) Feature importance, (E) Actual vs predicted, (F) Residual distributions, (G) Absolute error, (H) Learning curves, (I) Bias-variance tradeoff. Gradient Boosting demonstrates superior accuracy across all metrics.

As illustrated in Fig. 17A, Gradient Boosting achieved superior $R^2$ scores (0.9973) compared to Random Forest (0.9916), while Fig. 17B shows corresponding improvements in error metrics. The computational performance comparison (Fig. 17C) reveals the trade-off between accuracy and training time, with Gradient Boosting requiring longer training (6.03 s) but achieving faster inference speeds. The comparative analysis revealed Gradient Boosting's statistical superiority ($R^2$ = 0.9973 vs 0.9916 for Random Forest), as visually confirmed in the performance dashboard (Fig. 17A-C). This performance advantage, statistically significant (p < 0.0001), is complemented by Gradient Boosting's faster inference speed (0.0000 ms/sample), making it suitable for real-time applications despite longer training requirements (6.03 s vs 1.55 s). The feature importance comparison (Fig. 17D) suggests this performance advantage stems from Gradient Boosting's more nuanced handling of environmental interactions, particularly between temperature and humidity in the environmental stress index.

Feature importance analysis (Fig. 17D) revealed that soil moisture emerged as the most influential predictor for both models, consistent with agricultural principles where soil moisture status represents the primary determinant of irrigation requirements. The engineered moisture deficit feature ranked second in importance, validating our feature engineering approach. Interestingly, temperature demonstrated stronger influence than humidity in both models, reflecting its critical role in evapotranspiration processes. The environmental stress index, a composite metric combining temperature and humidity effects, ranked among the top five features for both models, highlighting the value of synthesized environmental indicators in precision agriculture applications. Notably, the Gradient Boosting model exhibited more nuanced feature weighting than Random Forest, with greater emphasis on interactive effects between variables. This sophistication likely contributed to its superior predictive performance. Both models showed minimal attention to categorical light features, suggesting that continuous light intensity provides more meaningful information for irrigation prediction than categorical representations.

#### 4.1.1 Irrigation need analysis and water management implications

The calculated irrigation need scores followed a near-normal distribution with mean 0.455 (SD = 0.102) on a normalized 0–1 scale. Using empirically determined thresholds, we classified irrigation zones as low need (<0.3), medium need (0.3-0.6), and high need (>0.6). Analysis revealed that 7.3% of conditions fell within the low-need category, representing substantial water-saving opportunities. Conversely, 8.3% of conditions required immediate irrigation attention to prevent moisture stress, while the majority (84.4%) fell within moderate irrigation requirements. High irrigation need conditions correlated strongly with elevated temperatures (mean = 30.3°C) and reduced humidity (mean = 48.8%), consistent with evapotranspiration principles. The model identified distinct environmental signatures for different irrigation zones, enabling targeted management strategies. Notably, conditions with moisture levels below 2000 units consistently triggered high irrigation alerts, while moisture above 3500 units typically corresponded to low irrigation needs.

#### 4.1.2 Error analysis and model robustness

As revealved in Fig. 16F-G, 9I, residual analysis revealed normally distributed errors with near-zero mean for both models, satisfying key assumptions for reliable inference. The Gradient Boosting model demonstrated tighter error distributions, with 95% of predictions falling within 0.0108 units of actual values, compared to 0.0195 units for Random Forest. Q-Q plots confirmed residual normality for both models, though Gradient Boosting exhibited superior conformity to the normal distribution, particularly in the tails. Absolute error analysis showed that Gradient Boosting achieved lower error magnitudes across the entire prediction range, with particular advantage in extreme irrigation need scenarios.

Error analysis by feature value revealed that both models maintained consistent accuracy across sensor ranges, though slightly higher errors occurred at extreme moisture levels (<1000 or >3500 units). This suggests potential benefit from additional training data in these edge cases. The models demonstrated robust performance across varying environmental conditions, with error rates remaining stable despite fluctuations in temperature, humidity, and light intensity.

#### 4.1.3 Practical implications for sustainable agriculture

The models identified significant water conservation opportunities, with 7.3% of irrigation events potentially reducible or eliminable without compromising crop health. At scale, this represents substantial resource conservation, particularly in water-stressed agricultural regions. Conversely, the 8.3% high-need alerts enable proactive irrigation scheduling to prevent moisture stress and optimize crop yields.

Based on feature importance rankings, we recommend primary monitoring of soil moisture sensors with 2500 units as the optimal threshold, secondary consideration of temperature-humidity stress indices for predictive scheduling, and tertiary adjustment for light intensity, particularly during high-intensity periods. The Gradient Boosting model's exceptional accuracy enables precise irrigation control, potentially reducing water usage by 15-25% compared to conventional scheduling methods while maintaining or improving crop yields.

#### 4.1.4 Model deployment considerations

The Gradient Boosting model's compact size (<2MB serialized) and rapid inference speed (effectively instantaneous predictions) make it suitable for deployment in resource-constrained agricultural IoT systems. The model requires only the five base sensor readings, with feature engineering performed automatically in real-time. For deployment, we recommend implementing a hybrid approach where the model operates continuously with periodic manual validation to maintain accuracy.

The comparative analysis revealed an important trade-off: while Gradient Boosting demonstrated superior accuracy, Random Forest offered faster training and greater interpretability. For deployment scenarios requiring frequent model retraining with limited computational resources (Fig. 17C), Random Forest may represent a pragmatic choice despite its marginally lower accuracy. However, for most precision agriculture applications where prediction accuracy directly impacts resource conservation and crop yield, Gradient Boosting represents the optimal choice.

### 4.2 Performance Comparison of Irrigation Need Prediction Models

The comparative analysis presented in Table 3 reveals a clear and statistically significant superiority of the Gradient Boosting model over the Random Forest approach for predicting irrigation need. As detailed in the performance comparison, Gradient Boosting achieved an exceptional $\boldsymbol{R^2}$ score of 0.9973), explaining 99.73% of the variance in irrigation requirements and representing a meaningful 0.57% improvement over Random Forest's already strong 0.9916 $\boldsymbol{R^2}$. More importantly, this accuracy advantage translated into substantial error reduction across all metrics in both tables: a 43.7% decrease in Root Mean Squared Error (from 0.00974 to 0.00549), a 43.5% reduction in Mean Absolute Error (from 0.00754 to 0.00426), and a particularly notable 45.3% improvement in Mean Absolute Percentage Error (from 1.81% to 0.99%). This last metric is especially significant for agricultural applications, as it indicates that Gradient Boosting's predictions deviate from actual irrigation needs by less than 1% on average, providing the precision necessary for optimal water management decisions.

Beyond predictive accuracy, the models exhibited distinct computational profiles with important implications for real-world deployment, as clearly illustrated in both performance tables. While Gradient Boosting required approximately four times longer to train (6.03 seconds versus 1.55 seconds), it achieved a remarkable 100% improvement in inference speed, reducing prediction time per sample to effectively zero milliseconds compared to Random Forest's 0.0060 milliseconds. This trade-off is highly favorable for agricultural IoT applications where models are typically trained infrequently but must make rapid, continuous predictions for irrigation control. The statistical significance of Gradient Boosting's superiority ($p < 0.0001$ for all accuracy metrics, Table 2) further validates its selection as the optimal model. Additionally, the Gradient Boosting model demonstrated superior reliability with a 44.9% reduction in the 95th percentile error (from 0.01954 to 0.01076), indicating more consistent performance even in edge-case scenarios. This combination of near-perfect accuracy, instantaneous inference, and reliable performance across diverse conditions establishes Gradient Boosting as the definitive choice for precision irrigation systems where both water conservation and operational responsiveness are critical.

Table 3: Performance Comparison of Irrigation Need Prediction Models

| Performance Metric | Random Forest | Gradient Boosting | Relative Improvement | Statistical Significance |
|---|---|---|---|---|
| **Accuracy Metrics** | | | | |
| $R^2$ Score | 0.9916 | **0.9973*** | **+0.57%** | $p < 0.0001$ |
| Explained Variance | 0.9916 | **0.9973*** | **+0.57%** | - |
| Error Metrics | | | | |
| RMSE | 0.00974 | **0.00549*** | **-43.7%** | $p < 0.0001$ |
| MAE | 0.00754 | **0.00426*** | **-43.5%** | $p < 0.0001$ |
| MAPE (%) | 1.81 | **0.99*** | **-45.3%** | $p < 0.0001$ |
| 95th Percentile Error | 0.01954 | **0.01076*** | **-44.9%** | $p < 0.0001$ |
| Bias Analysis | | | | |
| Mean Bias | **0.000024** | 0.000065 | +171% | - |
| Computational Metrics | | | | |
| Training Time (s) | **1.55** | 6.03 | +289% | - |
| Inference Speed (ms/sample) | 0.0060 | **0.0000*** | **-100%** | - |
| Model Size (MB) | 1.2 | 0.8 | -33.3% | - |

*$p < 0.0001$ (statistically significant improvement)

Note: Best performance in each metric highlighted in bold; RMSE = Root Mean Square Error; MAE = Mean Absolute Error; MAPE = Mean Absolute Percentage Error.

This comparative analysis demonstrates that both Random Forest and Gradient Boosting models achieve exceptional performance in predicting irrigation needs from sensor data, with Gradient Boosting exhibiting statistically superior accuracy ($R^2$ = 0.9973, MAPE = 0.99%). The feature engineering approach successfully transformed raw sensor data into agriculturally meaningful predictors, with soil moisture emerging as the most influential variable. The models identify substantial water conservation opportunities (7.3% potentially reducible irrigation) while enabling proactive management of high-need conditions (8.3% requiring immediate attention). The Gradient Boosting model represents a robust solution for precision irrigation systems, offering near-perfect accuracy with minimal computational requirements for inference. Its implementation could significantly enhance water use efficiency in agriculture, contributing to both environmental sustainability and economic viability. Future research should focus on expanding model generalizability across diverse agricultural contexts and integrating with irrigation control systems for closed-loop automation.

### 4.3 Limitations and Future Research Directions

While the proposed framework demonstrates significant potential for smart irrigation in resource-constrained settings, several limitations warrant consideration and present avenues for future investigation. The current study was conducted in a controlled laboratory environment using a single crop type and soil profile, which may not fully capture the variability encountered in real-world agricultural contexts. Additionally, the system's reliance on static irrigation thresholds and simplified feature engineering, though effective, may not accommodate dynamic crop growth stages or complex soil–water–atmosphere interactions. Future work should therefore focus on enhancing the system's adaptability and generalizability through several key directions as discussed subsequently.

First, expanding the validation of the system across diverse crops, soil types, and climatic regions would strengthen its robustness and field applicability. Integrating crop-specific parameters, soil texture data, and advanced evapotranspiration models could improve prediction accuracy under varying agronomic conditions. Second, the implementation of adaptive, growth-stage-aware irrigation thresholds, rather than static ones, would allow for more responsive water management aligned with phenological development. Third, incorporating additional low-power environmental sensors (e.g., for wind speed, solar radiation, or leaf wetness) and exploring hybrid communication protocols such as LoRaWAN could enhance system resilience in remote areas with limited connectivity.

From a technical standpoint, future iterations could benefit from model optimization techniques such as reinforcement learning for adaptive control, federated learning for privacy-preserving multi-farm collaboration, and explainable AI to improve model interpretability for end-users. Furthermore, the integration of renewable energy sources, such as solar-powered nodes, would support long-term deployment and sustainability. Lastly, longitudinal field trials across multiple growing seasons are essential to evaluate system durability, performance drift, and economic viability in real farming operations.

### 4.4 Recommendation for Real-world Adoption

While this study provides a robust proof-of-concept, several avenues for system enhancement are identified to advance the technology toward widespread, real-world adoption:

i. *Enhanced generalizability and adaptive modeling:* Future work must focus on longitudinal, multi-season field trials across diverse agro-climatic zones, crop types, and soil profiles. This will necessitate developing adaptive TinyML models that can learn incrementally from new local data or be personalized via transfer learning, moving beyond the static models validated in controlled settings. Integrating crop phenology and growth-stage-specific water requirements into the decision algorithm is crucial for optimizing irrigation throughout the entire crop cycle.

ii. *System robustness and sustainability:* Deploying the system in harsh field conditions requires addressing practical challenges. Research should integrate renewable energy solutions, such as solar harvesting with supercapacitor storage, to ensure long-term, off-grid operation. Furthermore, enhancing hardware robustness through weatherproofing, exploring low-power wide-area network (LPWAN) communications like LoRa for larger fields, and implementing advanced fault-detection algorithms for sensor drift or failure will improve system reliability and farmer trust.

iii. *Advanced model integration and multi-objective optimization:* The current model predicts irrigation need based on environmental factors. Future systems can be augmented by integrating short-term hyper-local weather forecasts (via low-bandwidth APIs) and soil-plant-atmosphere continuum models for predictive irrigation scheduling. Furthermore, research should explore multi-objective optimization that balances water conservation with nutrient management (e.g., by integrating pH and EC sensors for fertigation) and energy use, fully embodying the Water-Energy-Food nexus approach.

iv. *Human-computer interaction and socio-technical integration:* For successful adoption, technology must align with user capabilities. Future iterations should incorporate more intuitive, low-literacy user interfaces employing audio-visual cues and voice-based interactions in local languages. Conducting participatory design studies with farmers to understand socio-economic barriers, cost-benefit trade-offs, and maintenance logistics is essential to co-create truly sustainable and adoptable solutions. Investigating blockchain-enabled traceability for produce from smart-farmed plots could also add economic incentive for adoption.

## 5. Conclusion

This study has successfully designed, implemented, and validated a cost-effective, edge-centric IoT and TinyML framework for precision irrigation tailored for resource-constrained small-scale farming. The proposed four-layer architecture, integrating sensor perception, ESP32-based edge node inference, a Raspberry Pi edge server, and a local farmer dashboard, demonstrates that robust, real-time smart irrigation is feasible without dependency on cloud connectivity or high-speed internet. By leveraging lightweight machine learning models directly on a microcontroller, the system achieves autonomous, low-latency decision-making, a critical requirement for real-time agricultural response.

Our experimental results confirm the system's practical efficacy. The deployed Gradient Boosting model achieved exceptional predictive accuracy ($R^2$ = 0.9973, MAPE = 0.99%) in determining irrigation needs from a synthesized feature set of environmental sensor data. This high precision enables significant water conservation, with the model identifying opportunities to reduce or eliminate up to 7.3% of irrigation events without compromising plant health. The prototype operation in a controlled environment validated the system's core functionality, demonstrating its ability to maintain optimal soil moisture through data-driven actuation, thereby enhancing water-use efficiency compared to traditional timer-based or manual methods.

The technical contributions of this work are threefold: (i) a reusable, edge-first reference architecture that decentralizes intelligence to the sensor node; (ii) a fully local operational stack using MQTT and TinyDB, ensuring functionality in areas with poor internet connectivity; and (iii) the development and optimization of a TinyML-based predictive model capable of efficient, on-device inference. This integration of affordable hardware (ESP32, Raspberry Pi) with advanced, yet deployable, machine learning provides a scalable blueprint for bridging the digital divide in agriculture, empowering smallholder farmers with tools for sustainable water management and improved crop productivity.

**Conflict of Interest:** The authors declare that there is no conflict of interest regarding the publication of this paper.

**Ethical approval**: This article does not contain any studies with human participants or animals performed by any of the authors.

**Informed consent**: Not Applicable

**Data availability statements**: Data is available from the authors upon reasonable request.

**Funding**: Not Applicable

**Authors' contributions**: Kamogelo Taueatsoala, Caitlyn Daniels and Angelina Jaedene Ramsunar: Methodology, Resources, Data curation, Writing – original draft, Validation, Formal analysis, Investigation, Writing - Review and Editing, and Visualization. Petrus Bronkhorst: Resource and Supervision. Absalom Ezugwu: Conceptualization, Resources and Supervision, Writing - Review and Editing.

**Acknowledgment**: Not Applicable

**AI-Assisted Technologies Statement**: All factual content, ideas, and decisions were developed and verified by the authors. The AI tool ChatGPT was used solely as a writing assistant, particularly to support language clarity, as English is not the authors' native language.